\documentclass{article}

\usepackage{array}
\usepackage[caption=false,font=normalsize,labelfont=sf,textfont=sf]{subfig}
\usepackage{textcomp}
\usepackage{stfloats}
\usepackage{url}
\usepackage{verbatim}
\usepackage{graphicx}
\usepackage{cite}
\usepackage{fullpage} 

\usepackage{adjustbox}

\newcommand{\xmark}{}

\usepackage{bm}
\usepackage{amsmath,amsfonts,amssymb,amsthm}

\newcommand{\multiset}[1]{
  \{\!\!\{#1\}\!\!\}
}

\newcommand*{\defeq}{\mathrel{\vcenter{\baselineskip0.5ex \lineskiplimit0pt
                     \hbox{\scriptsize.}\hbox{\scriptsize.}}}%
                     =}
\usepackage{algorithmic}
\usepackage{algorithm}
\usepackage[colorlinks=true,linkcolor=blue,citecolor=blue,allcolors=blue]{hyperref}
\usepackage[numbers, compress, sort]{natbib}
\usepackage{bbm}
\usepackage{booktabs}       
\usepackage{tcolorbox}
\usepackage{multirow}
\usepackage{caption}
\usepackage{mathtools}
\usepackage{nicefrac}

\usepackage{authblk}

\title{\textsc{In-n-Out}: Calibrating Graph Neural Networks \\for Link Prediction}

\author[1]{Erik Nascimento}
\author[2]{Diego Mesquita}  
\author[3,4]{Samuel Kaski} 
\author[1,3]{Amauri H. Souza}

\affil[1]{Federal Institute of Cear\'a}
\affil[2]{Getulio Vargas Foundation}
\affil[3]{Aalto University}
\affil[4]{University of Manchester}
\date{}

\begin{document}

\maketitle



\begin{abstract}
  Deep neural networks are notoriously miscalibrated, i.e., their outputs do not reflect the true probability of the event we aim to predict. While networks for tabular or image data are usually overconfident, recent works have shown that graph neural networks (GNNs) show the opposite behavior for node-level classification. But what happens when we are predicting links? We show that, in this case, GNNs often exhibit a mixed behavior. More specifically, they may be overconfident in negative predictions while being underconfident in positive ones. Based on this observation, we propose \texttt{IN-N-OUT}, the first-ever method to calibrate GNNs for link prediction. \texttt{IN-N-OUT} is based on two simple intuitions: i) attributing \emph{true/false} labels to an edge while respecting a GNNs prediction should cause but small fluctuations in that edge's embedding; and, conversely, ii) if we label that same edge contradicting our GNN, embeddings should change more substantially. An extensive experimental campaign shows that \texttt{IN-N-OUT} significantly improves the calibration of GNNs in link prediction, consistently outperforming the baselines available --- which are not designed for this specific task.\end{abstract}

\section{Introduction}
Graph neural networks \citep[GNNs,][]{scarselli2009,Gori2005} are powerful models that often interleave message-passing operators and non-linear activation functions to extract meaningful representations from relational data.
These methods have become the \emph{de facto} models for an array of sensitive tasks, including financial crime detection \citep{Cheng2020}, drug discovery \citep{antibiotic_design}, personalized medicine \citep{Wu2022}, and online advertising~\citep{advertising}.
As a consequence, domain experts may rely on GNN predictions to make critical decisions.
Given the potential risk behind these decisions, there is a growing concern regarding the poor calibration of GNNs --- a noticeable mismatch between confidence estimates and true label probabilities. 
To address this concern, prior works leverage graph structure to improve upon classical calibration methods, either by refining node-wise logits using a GNN \citep{CaGCN}, or using attention to aggregate logits from nearby nodes and compute node-wise temperature scales \citep{GATS}. 
Since these methods operate on node-level logits, they only apply to node classification.
However, to the best of our knowledge, there are no works dedicated to the calibration of GNNs for link prediction. 




A natural question ensues: 
\emph{can we use off-the-shelf solutions to calibrate GNNs for link prediction?} 
In principle, there are many methods designed for independently distributed data that we can use to alleviate miscalibration in GNNs --- e.g., isotonic regression \citep{isotonicRegression} and temperature scaling \citep{temperatureScalling}. These methods work directly in logit space, e.g., scaling them by a constant or mapping logits intervals to a sequence of increasing values. The downside of these approaches is that the logits are the final bottleneck in a neural network, and a great deal of information (possibly useful for calibration) may be lost along the way. 

In this paper, we propose \texttt{IN-N-OUT}, the first method for \emph{post-hoc} calibration of GNNs in the context of link prediction. \texttt{IN-N-OUT} is a novel temperature-scaling method that builds on a simple intuition: if a GNN believes an edge does (not) occur, (not) seeing it should not drastically change the information we have on the graph --- which is condensed in the GNN embeddings. On the other hand, if the GNN believes an edge does (not) occur, seeing the opposite should significantly change that edge's embedding. To encapsulate this concept, \texttt{IN-N-OUT} parameterizes the scaling factor as a function of how much an edge embedding changes by possibly observing an edge, modulated by the class predicted by the GNN.

Our experimental campaign, including multiple datasets and GNN models, shows that \texttt{IN-N-OUT} generally outperforms off-the-shelf calibration methods, consistently resulting in smaller expected calibration errors. 
More specifically, \texttt{IN-N-OUT} outperforms Isotonic Regression \citep{isotonicRegression}, Histogram binning \citep{histogramBinning}, Temperature scaling and BBQ \citep{BBQ}, producing lower calibration error on  29 out of 35 experiments. We also run an extensive ablation study with 4 GNNs and 8 datasets to validate the intuitions behind our modeling choices.

In summary, our \textbf{contributions} are:
\begin{enumerate}
    \item We show for the first time that GNNs are usually miscalibrated for link prediction tasks. While this phenomenon also happens for node classification, the calibration curves for link prediction are often more complex, exhibiting a mixed behavior: positive predictions being underconfident, and negative ones being overconfident;
    \item We propose \texttt{IN-N-OUT}, the first method for calibration of GNNs in the context of link prediction.
    \item Our experiments on real data show that \texttt{IN-N-OUT} consistently outperforms the existing baselines, which include popular calibration methods for non-relational data. We also perform an ablation study to assess the impact of the model components.
\end{enumerate}

\section{Background}

\vspace{6pt}\noindent\textbf{Notation.} We define a graph $\mathcal{G} = (V, E)$, with a set of nodes  $V = \{1, \ldots, n\}$ and a set of edges $E \subseteq V \times V$. We denote the adjacency matrix of $\mathcal{G}$ by $\bm{A} \in \mathbb{R}^{n\times n}$, i.e., $A_{ij}$ is one if $(i,j) \in E$ and zero otherwise. Let $\bm{D}$ be the diagonal degree matrix of $\mathcal{G}$, i.e., $D_{i i} = \sum_{j} {A}_{i j}$. We also define the \emph{normalized} adjacency matrix with added self-loops as $\widetilde{\bm{A}} = (\bm{D} + \bm{I}_n)^{-1/2} (\bm{A} + \bm{I}_n) (\bm{D} + \bm{I}_n)^{-1/2}$, where $\bm{I}_n$ is the $n$-dimensional identity matrix. Furthermore, let  $\bm{X} \in \mathbb{R}^{n \times d}$ be a matrix of $d$-dimensional node features. Therefore, throughout this work, we often represent a graph $\mathcal{G}$ using the pair $(\bm{A}, \bm{X})$, where $\bm{X} \in \mathbb{R}^{n \times d}$ comprises $\bm{x}_v$ in its $v$-th row. We also denote a node $v$'s neighborhood in $G$ by $\mathcal{N}_v$ and its degree by $d_v = |\mathcal{N}_v|$. Furthermore, we use $\multiset{\cdot}$ to denote multisets.

\vspace{6pt}\noindent\textbf{Graph neural networks.}  GNNs are neural networks that exploit a graph's structure to propagate information between its nodes to create meaningful embeddings, which encapsulate the information regarding their local topologies. One of the most popular ways to see them is through the lens of message-passing~\citep{Gilmer2017}. Within this framework, a GNN starts initializing each node's embedding with its original features, i.e., $\bm{h}^{(0)}_v = \bm{x}_v$ for all node $ v \in V$. At each layer $\ell$, in parallel, each node $v$ gathers messages $\bm{m}^{(\ell)}_{u \rightarrow v}$ from all its neighbors $u \in \mathcal{N}_v$, compiling them into an aggregated message $\bm{m}^{(\ell)}_v$:
\begin{equation*}
    \bm{m}_v^{(\ell)} = \texttt{Aggregate}\left(\multiset{ \bm{h}^{(\ell-1)}_u  : u \in \mathcal{N}_v}  \right),
\end{equation*}
where $\texttt{Aggregate}$ is a function defined on multisets, i.e., it is order-invariant. Subsequently, each node uses the (so-called) $\texttt{Update}$ function to refresh its embedding in light of the aggregated message:
\begin{equation*}
    \bm{h}_v^{(\ell)} = \texttt{Update}\left( \bm{m}_v^{(\ell)}, \bm{h}_v^{(\ell-1)}  \right).
\end{equation*}

 Notably, the majority of GNNs can be described in terms of message-passing by choosing an adequate pair of $\texttt{Aggregate}$/$\texttt{Update}$ functions, modifying $G$'s adjacency matrix, or incrementing node features~\citep{alltheway}. This is the case for the five GNN architectures we use in our experiments: VGAE, SAGE, PEG, GCN, GIN and SEAL. For instance, VGAE~\citep{VGAE} consists of a GCN \citep{GCN} trained in a variational auto-encoding framework~\citep{vae} with a dot product decoder. SAGE~\citep{SAGE} is a scalable GNN that randomly subsamples neighbors to speed up message passing, yielding noisy updates. 
 GIN \citep{xu2018gin} leverages injective aggregate/update functions to boost the expressive power of GNNs.
 PEG~\citep{PEG} augments node features with a positional embedding, which can capture topological information that cannot be assessed through message-passing alone. %
 SEAL~\citep{SEAL} outlines subgraphs to compute edge embeddings via message-passing, further annotating nodes via the \emph{double radius node labeling} trick to increase expressiveness. 
 For a more thorough overview of GNNs, see \citet{book-graph-learning}.

\vspace{6pt}\noindent\textbf{Link prediction.} In the context of link prediction, we can combine node embeddings after an arbitrary number of message-passing layers (say $L$) to build edge embeddings, which can be fed to an MLP to classify the respective edge. More specifically, we extract build and embedding $\bm{h}_{uv}$ for an edge $(u,v) \in V^2$ by applying some function $\psi$ to $\bm{h}_u^{(L)}$ and $\bm{h}_v^{(L)}$, i.e., $\bm{h}_{uv} = \psi(\bm{h}_u^{(L)},\bm{h}_v^{(L)})$. When we are dealing with an undirected edge, $\psi$ must also be order-invariant, otherwise, $\bm{h}_{uv} \neq \bm{h}_{vu}$. In general, we can write the probability for an edge (inferred by a GNN) as $p_{uv} = \texttt{MLP}(\bm{h}_{uv})$. 
A common metric for evaluating link prediction tasks is hits@k, i.e., \% of positive edges receiving higher logits than the negative edge w/ the kth highest logit.

\vspace{6pt}\noindent\textbf{Uncertainty/confidence calibration.} Consider a binary classification problem. Let $Y$ and $X$ be the random variables corresponding to the response (binary) variable and the input feature vector. Let also $h$ be a neural network and $h(\bm{x}) \in [0,1]$ be its output probability for the positive class. In this case, $\hat{P} \defeq h(X)$ also characterizes a random variable. We say $h$ is perfectly calibrated if
\begin{align}
    \mathbb{P}( Y = 1| \hat{P} = c) = c \quad \forall c \in [0,1]. 
    \label{calibrated}
\end{align}
If we have $M$ input vectors $\bm{x}_1, \bm{x}_2, \ldots, \bm{x}_M$ for which $h$ outputs the probability $c$, \autoref{calibrated} entails we expect $c M$ of those vectors to belong to class $1$ --- and we also expect $(1-c)M$ to belong to class $0$. It is well-known, however, that modern neural networks are poorly calibrated~\citep{temperatureScalling}. In the following, we review i) the \emph{expected calibration error}, which is the main metric used to evaluate calibration quality; ii) \emph{reliability diagrams}, which we can use to visually interpret the expected calibration error in deeper detail; and iii) \emph{temperature scaling}, one of the most popular methods for post-hoc calibration of modern neural networks.

\vspace{6pt}\noindent\textbf{The expected calibration error (ECE)} measures how far a model is from satisfying \autoref{calibrated}. To compute the ECE of a binary predictive model $h$ on a sequence of input vectors $\bm{x}_{1}, \ldots, \bm{x}_{M}$, we first compute the  probabilities $c_{1}\, \ldots, c_M$ for class $1$, with $c_m \defeq h(\bm{x}_m)$. For simplicity of presentation, we define the sequence $(c_1, y_1),\ldots, (c_M, y_M)$ of predicted probabilities/true label pairs. Finally, we divide this sequence into $N$ contiguous splits (i.e., ordered bins) $B_1,\ldots,B_N$ and compute the ECE as:
\begin{eqnarray}
    \text{ECE} = \frac{1}{M} \sum_{n=1}^{N} \left| \sum_{(c, y) \in B_n} (y - c) \right|
    \label{eq:ece}.
\end{eqnarray}

\vspace{6pt}\noindent\textbf{Reliability diagram.} Reliability Diagrams are plots that allow for visualizing the calibration as a function of $h$'s output probability~\citep{niculescu}. The diagram comprises a point for each bin $B_1,\ldots,B_N$ used to compute the ECE. Each point's horizontal coordinate is the average confidence within that bin, while the vertical one is the frequency of positive labels in that bin. A perfectly calibrated model should lie on the identity line, plotted with the reliability curve for reference.

\vspace{6pt}\noindent\textbf{Temperature-based calibration.} The main idea behind temperature-scaling consists of simply scaling all the logits from $h$ by $1/T$ for some $T>0$. More concretely, suppose we are dealing witha binary classification problem (e.g., link prediction) and let $g(\bm{x})$ denote the outputs of $h$ before being projected to the probability simplex (such that $h = \sigma \circ g$). Temperature scaling defines a corrected version of the output as $\hat{h}(\bm{x}) = \sigma(g(\bm{x}))$ for all input vector $\bm{x}$. Note that when $T\rightarrow\infty$, all predictions tend to $\hat{h}(\bm{x}) \rightarrow 1/2$ for any $\bm{x}$. On the other hand,  if $T\rightarrow 0$, $g(x)>0$ implies $\hat{h}(\bm{x}) \rightarrow 1$, and $g(x)<0$ implies $\hat{h}(\bm{x}) \rightarrow 0$. Put simply, small $T$ tends to decrease the entropy of our predictive distribution. Conversely, large $T$ tends to flatten our predictive, increasing entropy.
In this binary context, temperature scaling can be seen as a special case of Platt's scaling. The main difference between these methods is that the former does not add a bias to the logits --- in order to guarantee $h$'s accuracy is preserved. Importantly, $T$ is usually learned by minimizing the \emph{negative log likelihood} (NLL) on a labeled validation set, which was not seen during the training of the neural network $h$.

\begin{figure*}[th!]
\centering
\includegraphics[width=0.9\linewidth]{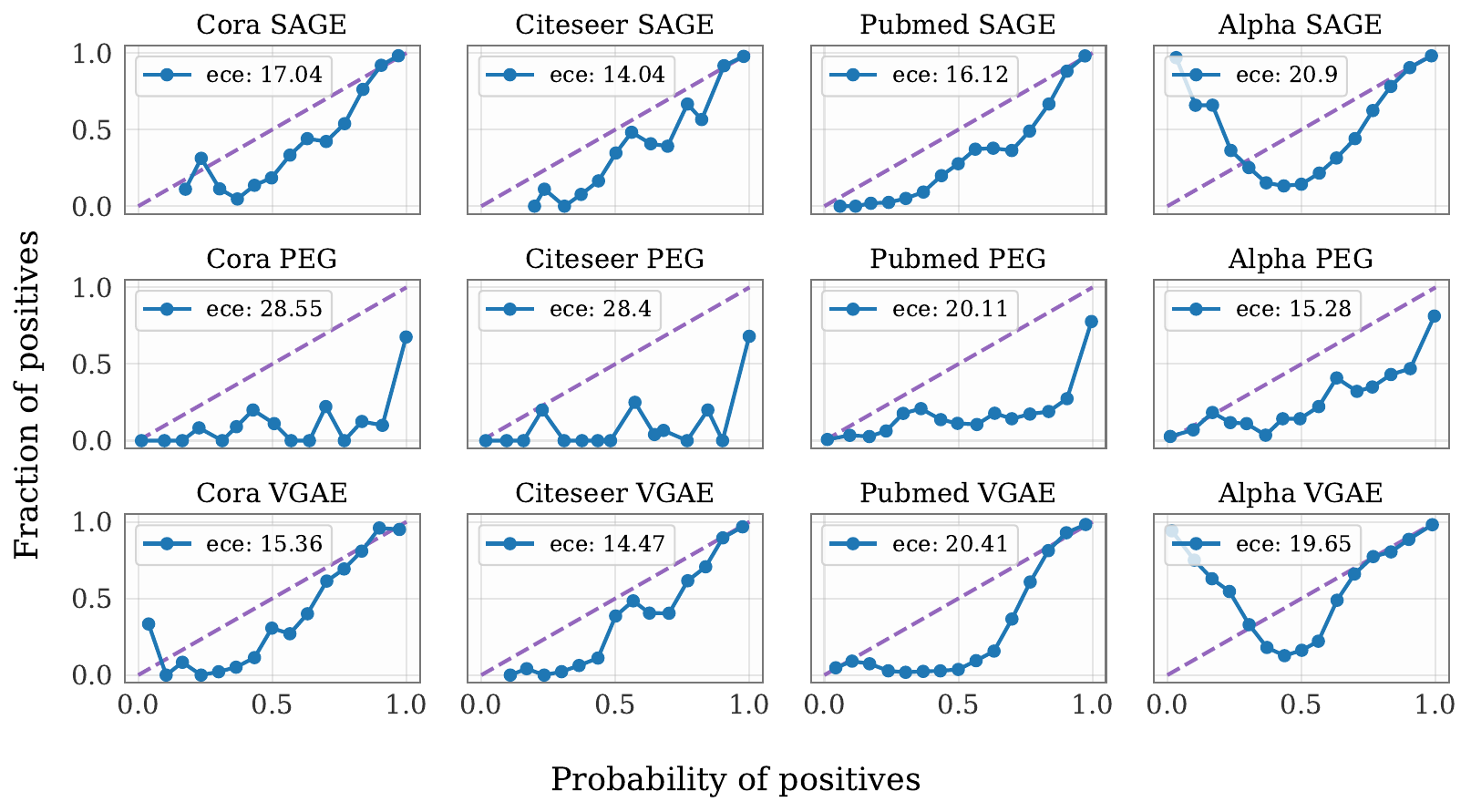}
 \caption{Reliability diagrams of VGAE, PEG and SAGE  trained on the cora, citeseer, and pubmed datasets. The dashed line is a baseline for perfect calibration. In most cases, the GNNs tend to attribute higher probability than they should to the positive class. On one hand, this hints at over-confident predictions for edges positively classified. On the other, this indicates under-confidence in our negative predictions. This is behavior is notably more complex than what was previously observed for GNN-based node classification~\citep{CaGCN, GATS}, in which predictions are overall underconfident.}
 \label{fig:overcofident}
\end{figure*}

\section{GNNs' link predictions are miscalibrated}\label{sec:calibration}

To assess the extent to which GNNs are miscalibrated, we consider three representative models and four datasets. More specifically, we compute reliability diagrams for PEG, VGAE, and SAGE  on the cora, citesser, pubMed, and bitcoin\_alpha datasets \citep{bitcoinALPHA}. Since we do not aim to benchmark these GNNs, we follow implementation guidelines from the original works, using the same training procedures and hyperparameters whenever available. Appendix A provides further details on implementation and datasets. 

\autoref{fig:overcofident} shows the reliability diagrams for the representative GNNs. Note that, in most cases, GNNs tend to be overconfident towards the positive class, assigning higher probability to the occurrence of a link than they should. For instance, for pubmed, less than $5\%$ of the links that VGAE predicts with probability $\approx$ 60\% truly occurs. For the cases when the GNNs output probabilities $<0.5$, this means that improving calibration implies increasing the probability for the negative class. Put briefly, GNNs are overconfident when they classify links as positives and underconfident otherwise. 
This pattern repeats in $\approx58\%$ of all combinations of dataset/GNN we experimented with.
Interestingly, this suggests that GNN-based link prediction results in more complex calibration pattern than observed for both non-relational deep neural networks~\citep{temperatureScalling}, which are usually overconfident; and GNN-based node classification~\citep{CaGCN,GATS}, which are usually underconfident. For completeness, we report reliability diagrams for additional datasets and GNNs architectures in the supplementary material.
It is also worth mentioning that the second most common pattern is U-shaped diagrams ($\approx25\%$), which denote mostly overconfidence in both negative and positive predictions.

\section{Calibration using \texttt{IN-N-OUT}}

When we are dealing with link prediction, GNNs are usually trained to distinguish nodes that are present in the input graph from spurious ones, created via negative sampling. Tipically, we propagate edge embeddings through a feedforward network to compute class probabilities. In this sense, these embeddings summarize all information the GNN deems relevant to infer if the endpoints will connect or not. In summary, the edge embeddings summarize all our knowledge regarding the relationships between the nodes. 
In light of this intuition, if our GNN points to two nodes connecting with high probability, observing this edge should not change our knowledge about it dramatically, i.e. the edge embedding should not vary substantially. On the other hand, if the GNN outputs a low probability for an edge, seeing it should cause perplexity, substantially changing the corresponding edge embedding. 
To capture this insight, we propose a temperature scale-based calibration method which we call \texttt{IN-N-OUT}. \texttt{IN-N-OUT} parameterizes the temperature as a function of (a discrepancy between) the embeddings of the edge we want to predict computed with and without that same edge in the input graph.

For concreteness, suppose we want to predict whether an edge $(u, v)$ occurs. Assume further that ${A}_{uv} = 0$ and let the augmented adjacency matrix $\bm{A}^{+}$ be defined as:
\begin{equation*}
    A^{+}_{ij} = \begin{cases}
        A_{ij}, & \text{ if } (i,j) \neq (u, v)\\
        1, & \text{ otherwise.}
    \end{cases}
\end{equation*}

Let $\bm{H} = \mathrm{GNN}(\bm{X}, \bm{A})$ such that their $u$-th row $\bm{h}_u$ is the embedding of node $u$. Following the standard practice from the GNN literature, we define the embedding of an edge $(u, v)$ aggregating the node embeddings $\bm{h}_u$ and $\bm{h}_v$ using a function $\psi$, i.e., $\bm{h}_{uv} = \psi(\bm{h}_u, \bm{h}_v)$. In the case where the edge $(u, v)$ is undirected, $\psi$ must is invariant to the order in which node embeddings are given, i.e., $\psi(\bm{h}_u, \bm{h}_v) = \psi(\bm{h}_v, \bm{h}_u)$. This order invariance is satisfied, for instance, by the dot product, the average, and the sum operator. To predict the occurrence of an edge $(u, v)$, we assume that the logit $s_{uv}$ is calculated using an MLP, i.e., $s_{uv} = \mathrm{MLP}(\bm{h}_{uv})$.

Then, let $\bm{H}^{+} = \mathrm{GNN}(\bm{X}, \bm{A}^{+})$ be the node embedding matrix given the incremented adjacency $\bm{A}^{+}$. Similarly to what we described earlier for the edge embedding $\bm{h}_{uv}$, let $\bm{h}^{+}_{uv} = \psi(\bm{h}^{+}_u, \bm{h}^{+}_v)$.
Intuitively, when $\bm{h}^{+}_{uv}$ and $\bm{h}_{uv}$ differ (in some sense) by a large margin, adding edge $(u,v)$ conflicts with the previous information that we had on the graph. 

In this sense, when our prediction $\hat{y}=\mathbbm{1}\left[s_{uv}>0\right]$ is positive, observing a wide margin between $\bm{h}^{+}_{uv}$ and $\bm{h}_{uv}$ should lower our confidence. 
On the other hand, when $\hat{y}=0$, the same observation reinforces the notion that the edge $(u,v)$ is \emph{alien} to the graph we are analyzing. Note that this reasoning is in line with the intuition we had regarding \autoref{fig:overcofident} --- i.e., that the over/under-confidence behavior in link-prediction changes depending on the most-likely class. With that in mind, we propose computing the temperature $T_{uv}$ as a function of a discrepancy measure $\gamma$ between the embeddings $\bm{h}_ {uv}$ and $\bm{h}^{+}_{uv}$:
\begin{equation*}
    T_{uv} = \begin{cases}
        \mathrm{MLP}_{c_1}\big(\gamma\left(\bm{h}_{uv}, \bm{h}^{+}_{uv}\right)\big) &  \text{ if }  s_{uv} > 0 \\
        \mathrm{MLP}_{c_2}\big(\gamma\left(\bm{h}_{uv}, \bm{h}^{+}_{uv}\right)\big) &  \text{ otherwise}
    \end{cases},
\end{equation*}
where $\mathrm{MLP}_{c_1}$ and $\mathrm{MLP}_{c_2}$ are multilayer perceptrons (MLPs) used for calibration, and return a scalar $T_{uv} \in \mathbb{R}^{+}$. To do so, the last layer can be equipped with a non-negative activation function such as the \textrm{Softplus}.  

We consider two options for $\gamma$: i) the Euclidean distance $\|\bm{h}_{uv} - \bm{h}^{+}_{uv}\|_2$, and ii) the difference $\bm{h}_{uv} - \bm{h}^{+}_{uv}$ between embeddings after/before edge inclusion.  
We treat this choice as a hyper-parameter.

To obtain a calibrated probability $\hat{p}_{uv}$ for an edge $(u,v)$, the \textit{logit} $s_{uv}$ from the original GNN is divided by the temperature $T_{uv}$ and passed through a sigmoid function, so that $\hat{p}_{uv}=\sigma(s_{uv} / T_{uv})$. 
Importantly, choosing the calibration network ($c_1$ or $c_2$) based on $s_{uv}$'s sign (which encodes whether the prediction is positive/negative) captures well the insight provided in Section \ref{sec:calibration}. More specifically, in the majority of cases, GNN models present opposite calibration behaviors for the positively and negatively predicted links. Nonetheless, our formulation is very general and can accommodate other minority patterns. 

\vspace{6pt}\noindent\textbf{Training procedure.} To apply $\texttt{IN-N-OUT}$, we are left with the task of learning $ \mathrm{MLP}_c$. We propose doing so by minimizing a conical combination of i) the NLL on a calibration set ($\mathcal{L}_{\mathrm{NLL}}$), comprising positive edges from the GNN's training set and negative ones obtained via negative sampling; ii) an ECE term ($\mathcal{L}_{\mathrm{ECE}}$), which directly penalizes miscalibration of our temperature-scaled model; and  iii) a penalty term that enforces low confidence on wrong predictions (most likely class $\neq$ ground truth) and high confidence on correct ones ($\mathcal{L}_{\mathrm{Cal}}$).   For concreteness, let $\mathcal{C} \subseteq V^2 \times \{0, 1\}$ be our calibration set. More specifically, $\mathcal{C}$ comprises triplets resulting from the concatenation of an edge and its respective label --- $1$ if observed and $0$ if obtained via negative sampling. Then, we can define the negative log-likelihood term as:
\begin{alignat}{2}
    \mathcal{L}_{\mathrm{NLL}} &= \sum_{(u,v, y) \in \mathcal{C}}  - y \log (\hat{p}_{uv}) - (1-y)\log (1 - \hat{p}_{uv})
    \label{eq:nll},
\end{alignat}
$\mathcal{L}_{\mathrm{ECE}}$ is defined according to \autoref{eq:ece} using fifteen equally-spaced bins, and we define $\mathcal{L}_{\mathrm{Cal}}$ as: 
\begin{align}
    \mathcal{L}_{\mathrm{Cal}} &= \frac{1}{|\mathcal{C}|}  \sum_{(u,v,y) \in \mathcal{C}} - (2y-1) \hat{p}_{uv} \label{eq:lcal}
\end{align}

Recall that temperature scaling does not change the most likely classes, so \autoref{eq:lcal} drives the probability $\hat{p}_{uv}$ to $0.5$ if the GNN classifies the example incorrectly. Otherwise, if $y=1$ it pushes $\hat{p}_{uv}$ towards one; and if $y=0$, it drives $\hat{p}_{uv}$ towards zero.
Also, $\mathcal{L}_{\mathrm{Cal}}$ can be seen as a simplified version (for binary outcomes) of the calibration penalty proposed by \citet{CaGCN} in the context of node classification. 

To summarize, our loss function is:
\begin{align}
    \mathcal{L} = \mathcal{L}_{\mathrm{NLL}} + \mathcal{L}_{\mathrm{Cal}} + \lambda \, \mathcal{L}_{\mathrm{ECE}},
\end{align}
where $\lambda>0$ is a hyperparameter that we choose by minimizing the ECE in a validation set.
It is worth mentioning that we sample negative edges by selecting uniformly at random a pair of nodes for which an edge was not previously observed.

\section{Related works}

While there are no prior works in the context of link prediction, a few works have tackled the issue for GNN-based node classification. \citet{brunoMiscalibration} first pointed to the issue of miscalibration of GNNs. \citet{CaGCN} highlighted an overall underconfident behavior and proposed CaGCN, a method that leverages node embeddings from auxiliary GNN temperature scales, outperforming off-the-shelf calibration strategies. More recently, \citet{GATS} further analyzed factors that affect GNN calibration (e.g., diversity of node-wise predictive distributions, and neighborhood similarity) on node classification and devised an attention-based temperature scaling method to address them directly. 
\citet{Wang22_GCL} argued that the shallow nature of GNNs is the culprit behind their underconfidence and proposed counter-acting it using a minimal-entropy regularization term. 
\citet{ConformalGNN} and \citet{huang2023uncertainty} proposed methods to compute conformal intervals for GNNs. Furthermore, \citet{vos2023calibration} analyzed the calibration of GNNs within the context of medical imaging tasks.

\begin{table*}[ht!]
\centering
\caption{Expected calibration error (mean and std deviation). The best ECE results found for each GNN are highlighted in \textcolor{blue}{blue} (the lower the value, the better the model). Notably, \texttt{IN-N-OUT} outperforms the baselines in $\approx83\%$ of the cases.}  

\adjustbox{width=\textwidth}{
\begin{tabular}{llrrrrrrrr}
\toprule
\color{black} & \textbf{Model} \color{black} & \textbf{ Uncalibrated} \color{black} & \textbf{Iso} \color{black} & \textbf{ Temp} \color{black} & \textbf{ BBQ} \color{black} & \textbf{ Hist} \color{black} & \textbf{ IN-N-OUT}\\ 
\toprule

 \parbox[t]{5mm}{\multirow{6}{*}{\rotatebox[origin=c]{0}{Cora}}}
 \color{black} & VGAE \color{black} & 15.02 \color{gray} $\pm$ 0.76 \color{black} & 6.88 \color{gray} $\pm$ 1.72 \color{black} & 12.62 \color{gray} $\pm$ 1.79 \color{black} & 7.09 \color{gray} $\pm$ 1.09 \color{black} & 6.14 \color{gray} $\pm$ 0.76 \color{black} & \textbf{ \color{blue}3.33} \color{gray} $\pm$ 0.67\\
 
 \color{black} & SAGE \color{black} & 17.23 \color{gray} $\pm$ 1.74 \color{black} & 17.36 \color{gray} $\pm$ 1.84 \color{black} & 17.20 \color{gray} $\pm$ 1.47 \color{black} & 17.66 \color{gray} $\pm$ 1.10 \color{black} & 17.61 \color{gray} $\pm$ 1.19 \color{black} & \textbf{ \color{blue}6.84} \color{gray} $\pm$ 0.48\\
 
 \color{black} & PEG \color{black} & 27.91 \color{gray} $\pm$ 0.60 \color{black} & 27.65 \color{gray} $\pm$ 1.34 \color{black} & 23.90 \color{gray} $\pm$ 1.55 \color{black} & 27.30 \color{gray} $\pm$ 1.40 \color{black} & 27.43 \color{gray} $\pm$ 1.89  \color{black} & \textbf{ \color{blue}11.53} \color{gray} $\pm$ 1.40\\
 
 \color{black} & GIN \color{black} & 15.07 \color{gray} $\pm$ 1.29 \color{black} & 16.66 \color{gray} $\pm$ 1.62 \color{black} & 14.23 \color{gray} $\pm$ 1.63 \color{black} & 16.36 \color{gray} $\pm$ 1.90 \color{black} & 16.63 \color{gray} $\pm$ 1.87 \color{black} & \textbf{ \color{blue}4.05} \color{gray} $\pm$ 1.31\\

 \color{black} & GCN \color{black} & 18.34 \color{gray} $\pm$ 1.13 \color{black} & 12.62 \color{gray} $\pm$ 0.97 \color{black} & 15.96 \color{gray} $\pm$ 1.11 \color{black} & 12.13 \color{gray} $\pm$ 1.30 \color{black} & 12.62 \color{gray} $\pm$ 1.08 \color{black} & \textbf{ \color{blue}6.90} \color{gray} $\pm$ 0.79\\

 \color{black} & SEAL \color{black} & 4.89 \color{gray} $\pm$ 0.18 \color{black} & 3.27 \color{gray} $\pm$ 0.15 \color{black} & 3.99 \color{gray} $\pm$ 0.66 \color{black} & 3.63 \color{gray} $\pm$ 0.24 \color{black} & 3.58 \color{gray} $\pm$ 0.20 \color{black} & \textbf{ \color{blue}2.41} \color{gray} $\pm$ 0.36\\
 
\midrule

 \parbox[t]{5mm}{\multirow{6}{*}{\rotatebox[origin=c]{0}{Cite}}}
 \color{black} & VGAE \color{black} & 14.00 \color{gray} $\pm$ 0.67 \color{black} & 5.52 \color{gray} $\pm$ 1.81 \color{black} & 18.57 \color{gray} $\pm$ 1.86 \color{black} & 5.14 \color{gray} $\pm$ 1.29 \color{black} & 5.02 \color{gray} $\pm$ 1.36 \color{black} & \textbf{ \color{blue}4.19} \color{gray} $\pm$ 0.65\\
 
 \color{black} & SAGE \color{black} & 14.52 \color{gray} $\pm$ 1.89 \color{black} & 18.81 \color{gray} $\pm$ 1.14 \color{black} & 12.86 \color{gray} $\pm$ 1.56 \color{black} & 18.64 \color{gray} $\pm$ 1.90 \color{black} & 17.88 \color{gray} $\pm$ 1.82 \color{black} & \textbf{ \color{blue}3.79} \color{gray} $\pm$ 1.98\\
 
 \color{black} & PEG \color{black} & 28.39 \color{gray} $\pm$ 1.59 \color{black} & 25.15 \color{gray} $\pm$ 1.74 \color{black} & 22.30 \color{gray} $\pm$ 1.35 \color{black} & 24.82 \color{gray} $\pm$ 1.76 \color{black} & 24.93 \color{gray} $\pm$ 0.56 \color{black} & \textbf{ \color{blue}12.02} \color{gray} $\pm$ 0.98\\
 
 \color{black} & GIN \color{black} & 15.69 \color{gray} $\pm$ 1.89 \color{black} & 23.86 \color{gray} $\pm$ 2.96 \color{black} & 13.98 \color{gray} $\pm$ 1.96 \color{black} & 23.80 \color{gray} $\pm$ 2.65 \color{black} & 23.30 \color{gray} $\pm$ 2.09 \color{black} & \textbf{ \color{blue}2.20} \color{gray} $\pm$ 0.86\\

 \color{black} & GCN \color{black} & 16.84 \color{gray} $\pm$ 1.62 \color{black} & 12.55 \color{gray} $\pm$ 1.43 \color{black} & 14.37 \color{gray} $\pm$ 1.19 \color{black} & 11.98 \color{gray} $\pm$ 1.59 \color{black} & 10.80 \color{gray} $\pm$ 1.62 \color{black} & \textbf{ \color{blue}6.06} \color{gray} $\pm$ 1.23\\

   \color{black} & SEAL \color{black} & 4.22 \color{gray} $\pm$ 0.31 \color{black} & 3.59 \color{gray} $\pm$ 0.49 \color{black} & 3.67 \color{gray} $\pm$ 0.35 \color{black} & 3.62 \color{gray} $\pm$ 0.30 \color{black} & 3.50 \color{gray} $\pm$ 0.24 \color{black} & \textbf{ \color{blue}2.93} \color{gray} $\pm$ 0.28\\
   
\midrule

 \parbox[t]{5mm}{\multirow{6}{*}{\rotatebox[origin=c]{0}{Pub}}}
 \color{black} & VGAE \color{black} & 20.41 \color{gray} $\pm$ 0.64 \color{black} & 3.04 \color{gray} $\pm$ 1.52 \color{black} & 20.83 \color{gray} $\pm$ 1.86 \color{black} & 2.22 \color{gray} $\pm$ 1.59 \color{black} &  \color{black}2.20 \color{gray} $\pm$ 1.97 & \textbf{\color{blue}1.85} \color{gray} $\pm$ 0.37\\
 
 \color{black} & SAGE \color{black} & 18.78 \color{gray} $\pm$ 1.42 \color{black} & 4.93 \color{gray} $\pm$ 1.41 \color{black} & 13.95 \color{gray} $\pm$ 1.72 \color{black} & 4.95 \color{gray} $\pm$ 1.57 \color{black} & 4.83 \color{gray} $\pm$ 1.32 \color{black} & \textbf{ \color{blue}3.01} \color{gray} $\pm$ 1.32\\
 
\color{black} & PEG \color{black} & 20.49 \color{gray} $\pm$ 1.31 \color{black} & 20.31 \color{gray} $\pm$ 1.93 \color{black} & 20.34 \color{gray} $\pm$ 1.28 \color{black} & 20.22 \color{gray} $\pm$ 0.28 \color{black} & 20.00 \color{gray} $\pm$ 1.39 \color{black} & \textbf{ \color{blue}8.21} \color{gray} $\pm$ 0.75\\

 \color{black} & GIN \color{black} & 15.33 \color{gray} $\pm$ 1.34 \color{black} & 12.72 \color{gray} $\pm$ 1.30 \color{black} & 14.96 \color{gray} $\pm$ 1.08 \color{black} & 12.66 \color{gray} $\pm$ 1.34 \color{black} & 12.67 \color{gray} $\pm$ 1.12 \color{black} & \textbf{ \color{blue}3.73} \color{gray} $\pm$ 0.89\\

 \color{black} & GCN \color{black} & 20.98 \color{gray} $\pm$ 1.79 \color{black} & 3.54 \color{gray} $\pm$ 1.97 \color{black} & 9.13 \color{gray} $\pm$ 1.97 \color{black} & \textbf{ \color{blue}3.34} \color{gray} $\pm$ 1.68 \color{black} & 3.44 \color{gray} $\pm$ 1.75 \color{black} & 4.44 \color{gray} $\pm$ 1.46\\

 \color{black} & SEAL \color{black} & 4.32 \color{gray} $\pm$ 0.33 \color{black} & 4.10 \color{gray} $\pm$ 0.17 \color{black} & 4.04 \color{gray} $\pm$ 0.16 \color{black} & 3.82 \color{gray} $\pm$ 0.48 \color{black} & 4.13 \color{gray} $\pm$ 0.32 \color{black} & \textbf{ \color{blue}2.42} \color{gray} $\pm$ 0.34\\
   
\midrule

 \parbox[t]{5mm}{\multirow{5}{*}{\rotatebox[origin=c]{0}{Twitch}}}
 \color{black} & VGAE \color{black} & 16.32 \color{gray} $\pm$ 0.20 \color{black} & 1.84 \color{gray} $\pm$ 1.10 \color{black} & 16.17 \color{gray} $\pm$ 1.45 \color{black} & 1.72 \color{gray} $\pm$ 1.13 \color{black} & 1.66 \color{gray} $\pm$ 1.24 \color{black} & \textbf{ \color{blue}1.50} \color{gray} $\pm$ 0.44\\
 
 \color{black} & SAGE \color{black} & 16.84 \color{gray} $\pm$ 2.84 \color{black} & 2.17 \color{gray} $\pm$ 1.37 \color{black} & 23.96 \color{gray} $\pm$ 1.53 \color{black} & 2.08 \color{gray} $\pm$ 1.14 \color{black} & 1.90 \color{gray} $\pm$ 1.42 \color{black} & \textbf{ \color{blue}1.53} \color{gray} $\pm$ 0.89\\
 
 \color{black} & PEG \color{black} & 7.90 \color{gray} $\pm$ 0.91 \color{black} & 5.36 \color{gray} $\pm$ 1.19 \color{black} & 5.81 \color{gray} $\pm$ 1.19 \color{black} & 5.35 \color{gray} $\pm$ 1.11 \color{black} & 5.38 \color{gray} $\pm$ 0.09 \color{black} & \textbf{ \color{blue}2.21} \color{gray} $\pm$ 0.51\\

 \color{black} & GIN \color{black} & 17.86 \color{gray} $\pm$ 2.13 \color{black} & 1.91 \color{gray} $\pm$ 0.97 \color{black} & 13.78 \color{gray} $\pm$ 2.36 \color{black} & \textbf{\color{blue}1.85} \color{gray} $\pm$ 0.96 \color{black} & 1.86 \color{gray} $\pm$ 0.78 \color{black} & 2.69 \color{gray} $\pm$ 0.96\\

 \color{black} & GCN \color{black} & 16.09 \color{gray} $\pm$ 1.59 \color{black} & 2.56 \color{gray} $\pm$ 0.96 \color{black} & 6.79 \color{gray} $\pm$ 1.12 \color{black} & 1.66 \color{gray} $\pm$ 0.93 \color{black} & \textbf{\color{blue}1.61} \color{gray} $\pm$ 0.97 \color{black} & 4.16 \color{gray} $\pm$ 1.13\\

\midrule 

 \parbox[t]{10mm}{\multirow{6}{*}{\rotatebox[origin=c]{0}{Chame}}}
 \color{black} & VGAE \color{black} & 7.52 \color{gray} $\pm$ 1.63 \color{black} & 2.83 \color{gray} $\pm$ 1.26 \color{black} & 6.39 \color{gray} $\pm$ 1.46 \color{black} & 2.84 \color{gray} $\pm$ 1.22 \color{black} & 2.77 \color{gray} $\pm$ 1.29 \color{black} & \textbf{ \color{blue}1.07} \color{gray} $\pm$ 0.34\\
 
 \color{black} & SAGE \color{black} & 5.98 \color{gray} $\pm$ 3.61 \color{black} & 3.46 \color{gray} $\pm$ 1.44 \color{black} & 12.0 \color{gray} $\pm$ 1.64 \color{black} & 3.59 \color{gray} $\pm$ 1.76 \color{black} & 3.23 \color{gray} $\pm$ 1.56 \color{black} & \textbf{\color{blue}2.46} \color{gray} $\pm$ 0.86\\

 \color{black} & PEG \color{black} & 4.44 \color{gray} $\pm$ 0.24 \color{black} & 2.84 \color{gray} $\pm$ 0.39 \color{black} & 2.28 \color{gray} $\pm$ 0.13 \color{black} & 2.82 \color{gray} $\pm$ 0.18 \color{black} & 2.44 \color{gray} $\pm$ 0.32 \color{black} & \textbf{\color{blue}1.66} \color{gray} $\pm$ 0.36\\


 \color{black} & GIN \color{black} & 20.59 \color{gray} $\pm$ 1.96 \color{black} & 2.39 \color{gray} $\pm$ 1.23 \color{black} & 9.78 \color{gray} $\pm$ 1.08 \color{black} & 2.47 \color{gray} $\pm$ 1.17 \color{black} & 2.25 \color{gray} $\pm$ 1.31 \color{black} & \textbf{\color{blue}1.58} \color{gray} $\pm$ 0.97\\

 \color{black} & GCN \color{black} & 26.97 \color{gray} $\pm$ 1.94 \color{black} & 1.69 \color{gray} $\pm$ 0.96 \color{black} & 12.96 \color{gray} $\pm$ 1.63 \color{black} & \textbf{\color{blue}1.40} \color{gray} $\pm$ 0.97 \color{black} & 1.76 \color{gray} $\pm$ 0.95 \color{black} & 3.39 \color{gray} $\pm$ 0.92\\

 \color{black} & SEAL \color{black} & 3.21 \color{gray} $\pm$ 0.49 \color{black} & 2.67 \color{gray} $\pm$ 0.33 \color{black} & 2.64 \color{gray} $\pm$ 0.29 \color{black} & 3.31 \color{gray} $\pm$ 0.26 \color{black} & 2.98 \color{gray} $\pm$ 0.22 \color{black} & \textbf{ \color{blue}2.15} \color{gray} $\pm$ 0.32\\
 
\midrule

 \parbox[t]{7mm}{\multirow{6}{*}{\rotatebox[origin=c]{0}{Alpha}}}
 \color{black} & VGAE \color{black} & 20.36 \color{gray} $\pm$ 1.53 \color{black} & 9.42 \color{gray} $\pm$ 1.02 \color{black} & 10.89 \color{gray} $\pm$ 1.32 \color{black} & 8.71 \color{gray} $\pm$ 0.89 \color{black} & 8.65 \color{gray} $\pm$ 1.21 \color{black} & \textbf{\color{blue}8.42} \color{gray} $\pm$ 0.68\\
 
 \color{black} & SAGE \color{black} & 21.26 \color{gray} $\pm$ 1.37 \color{black} & 13.35 \color{gray} $\pm$ 1.02 \color{black} & 12.95 \color{gray} $\pm$ 1.31 \color{black} & 13.19 \color{gray} $\pm$ 1.13 \color{black} & 13.24 \color{gray} $\pm$ 1.40 \color{black} & \textbf{\color{blue}6.68} \color{gray} $\pm$ 0.91\\

 \color{black} & PEG \color{black} & 13.91 \color{gray} $\pm$ 1.13 \color{black} & 15.50 \color{gray} $\pm$ 1.09 \color{black} & 14.21 \color{gray} $\pm$ 1.16 \color{black} & 15.48 \color{gray} $\pm$ 2.02 \color{black} & 15.35 \color{gray} $\pm$ 1.03 \color{black} & \textbf{\color{blue}9.78} \color{gray} $\pm$ 0.89\\

 \color{black} & GIN \color{black} & 24.32 \color{gray} $\pm$ 2.63 \color{black} & 8.89 \color{gray} $\pm$ 1.12 \color{black} & 16.93 \color{gray} $\pm$ 1.63 \color{black} & 8.84 \color{gray} $\pm$ 1.36 \color{black} & 8.80 \color{gray} $\pm$ 1.17 \color{black} & \textbf{\color{blue}3.32} \color{gray} $\pm$ 0.89\\

 \color{black} & GCN \color{black} & 20.73 \color{gray} $\pm$ 1.75 \color{black} & 6.31 \color{gray} $\pm$ 1.97 \color{black} & 9.82 \color{gray} $\pm$ 1.19 \color{black} & 6.37 \color{gray} $\pm$ 1.58 \color{black} & 6.38 \color{gray} $\pm$ 1.67 \color{black} & \textbf{\color{blue}3.69} \color{gray} $\pm$ 1.98\\

   \color{black} & SEAL \color{black} & 2.60 \color{gray} $\pm$ 0.93 \color{black} & 3.68 \color{gray} $\pm$ 0.50 \color{black} & 3.25 \color{gray} $\pm$ 0.63 \color{black} & 3.41 \color{gray} $\pm$ 0.47 \color{black} & 3.39 \color{gray} $\pm$ 0.51 \color{black} & \textbf{\color{blue}2.04} \color{gray} $\pm$ 0.66\\
 
 \midrule

 \parbox[t]{7mm}{\multirow{6}{*}{\rotatebox[origin=c]{0}{OTC}}}
 \color{black} & VGAE \color{black} & 18.91 \color{gray} $\pm$ 1.62 \color{black} & 6.05 \color{gray} $\pm$ 1.19 \color{black} & 6.12 \color{gray} $\pm$ 0.98 \color{black} & 5.97 \color{gray} $\pm$ 0.69 \color{black} & \textbf{\color{blue}5.39} \color{gray} $\pm$ 0.87 \color{black} & 9.87 \color{gray} $\pm$ 0.97\\
 
 \color{black} & SAGE \color{black} & 24.04 \color{gray} $\pm$ 1.11 \color{black} & 13.00 \color{gray} $\pm$ 0.97 \color{black} & 14.58 \color{gray} $\pm$ 1.62 \color{black} & 12.86 \color{gray} $\pm$ 1.09 \color{black} & \textbf{\color{blue}12.68} \color{gray} $\pm$ 0.96 \color{black} & 15.09 \color{gray} $\pm$ 1.21\\

 \color{black} & PEG \color{black} & 12.89 \color{gray} $\pm$ 0.86 \color{black} & 11.94 \color{gray} $\pm$ 0.99 \color{black} & 9.62 \color{gray} $\pm$ 0.86 \color{black} & 12.03 \color{gray} $\pm$ 1.01 \color{black} & 12.09 \color{gray} $\pm$ 0.95 \color{black} & \textbf{\color{blue}6.35} \color{gray} $\pm$ 0.95\\
 
 \color{black} & GIN \color{black} & 20.01 \color{gray} $\pm$ 1.46 \color{black} & 2.55 \color{gray} $\pm$ 0.68 \color{black} & 12.96 \color{gray} $\pm$ 1.46 \color{black} & 2.08 \color{gray} $\pm$ 0.82 \color{black} & 2.02 \color{gray} $\pm$ 0.76 \color{black} & \textbf{\color{blue}1.23} \color{gray} $\pm$ 0.49\\

 \color{black} & GCN \color{black} & 20.55 \color{gray} $\pm$ 1.92 \color{black} & 5.46 \color{gray} $\pm$ 1.71 \color{black} & 10.34 \color{gray} $\pm$ 1.64 \color{black} & 5.34 \color{gray} $\pm$ 1.05 \color{black} & 5.42 \color{gray} $\pm$ 1.62 \color{black} & \textbf{\color{blue}4.88} \color{gray} $\pm$ 1.97\\

  \color{black} & SEAL \color{black} & 2.61 \color{gray} $\pm$ 0.89 \color{black} & 3.07 \color{gray} $\pm$ 0.62 \color{black} & 3.12 \color{gray} $\pm$ 0.74 \color{black} & 2.66 \color{gray} $\pm$ 0.70 \color{black} & 2.42 \color{gray} $\pm$ 0.62 \color{black} & \textbf{\color{blue} 2.19} \color{gray} $\pm$ 0.64\\

  \midrule

 \parbox[t]{7mm}{\multirow{2}{*}{\rotatebox[origin=c]{0}{DDI}}}
 \color{black} & VGAE \color{black} & 9.95 \color{gray} $\pm$ 1.24 \color{black} & 5.69 \color{gray} $\pm$ 0.63 \color{black} & 5.52 \color{gray} $\pm$ 0.75 \color{black} & 5.74 \color{gray} $\pm$ 0.90 \color{black} & 5.62 \color{gray} $\pm$ 0.81 \color{black} & \textbf{\color{blue}4.09} \color{gray} $\pm$ 0.99 \\

  \color{black} & PEG \color{black} & 9.04 \color{gray} $\pm$ 1.16 \color{black} & 6.39 \color{gray} $\pm$ 0.22 \color{black} & 6.59 \color{gray} $\pm$ 0.71 \color{black} & 6.54 \color{gray} $\pm$ 0.53 \color{black} & 6.51 \color{gray} $\pm$ 0.42 \color{black} & \textbf{\color{blue} 4.33} \color{gray} $\pm$ 0.48 \\

  \midrule

 \parbox[t]{7mm}{\multirow{2}{*}{\rotatebox[origin=c]{0}{Collab}}}
 \color{black} & VGAE \color{black} & 7.27 \color{gray} $\pm$ 0.73 \color{black} & 5.34 \color{gray} $\pm$ 0.43 \color{black} & 5.44 \color{gray} $\pm$ 0.52 \color{black} & 5.02 \color{gray} $\pm$ 0.49 \color{black} & 4.58 \color{gray} $\pm$ 0.62 \color{black} & \textbf{\color{blue}3.42} \color{gray} $\pm$ 0.58 \\

\color{black} & PEG \color{black} & 4.73 \color{gray} $\pm$ 0.81 \color{black} & 3.11 \color{gray} $\pm$ 0.52 \color{black} & 3.08 \color{gray} $\pm$ 0.56 \color{black} & 3.94 \color{gray} $\pm$ 0.30 \color{black} & 3.57 \color{gray} $\pm$ 0.46 \color{black} & \textbf{\color{blue}1.42} \color{gray} $\pm$ 0.35\\
\bottomrule
\end{tabular}
}
\label{tab:ece_results}
\end{table*}
\section{Experiments}
In this section, we assess the performance of \texttt{IN-N-OUT} on a variety of datasets and GNNs. 
Additionally, we run an ablation study to analyze the impact of our design choices on predictive performance. We implemented all experiments using PyTorch \citep{pytorch} and Torch Geometric \citep{torch_geometric}. Our code is attached as supplementary material.

\subsection{Evaluation setup}

\vspace{6pt}\noindent\textbf{Datasets.} We consider seven real-world datasets for link prediction: Cora, Citeseer, PubMed \citep{citationsDatasets}, Twitch, Chameleon \citep{twitchANDchameleon}, Bitcoin-Alpha and Bitcoin-OTC \citep{bitcoinALPHA,bitcoinOTC}. The first tree datasets are citation graphs in which nodes represent documents from different domains, and edges are citation links between these documents. The Twitch dataset comprises attributed graphs where nodes correspond to users, edges denote mutual friendships, and node attributes encode preferred games, locations, and streaming habits. Chameleon is a network of Wikipedia pages about chameleons. Its nodes represent articles, and edges reflect the mutual links between them. Node features encode the presence of specific nouns in articles and the average monthly traffic from October 2017 to November 2018.
Bitcoin-Alpha and Bitcoin-OTC are networks in which nodes correspond to user accounts trading Bitcoin. A directed edge $(u,v)$ denotes the degree of reliability assigned by user $u$ to user $v$, i.e., each edge has a score denoting the degree of trust. 
The ogbl-ddi dataset is a homogeneous, unweighted, undirected graph, representing the drug-drug interaction network. The ogbl-collab dataset is an undirected graph, representing a subset of the collaboration network between authors indexed by MAG.
Appendix shows summary statistics of the datasets.

\begin{figure*}[ht!]
\centering
\includegraphics[width=0.9\linewidth]{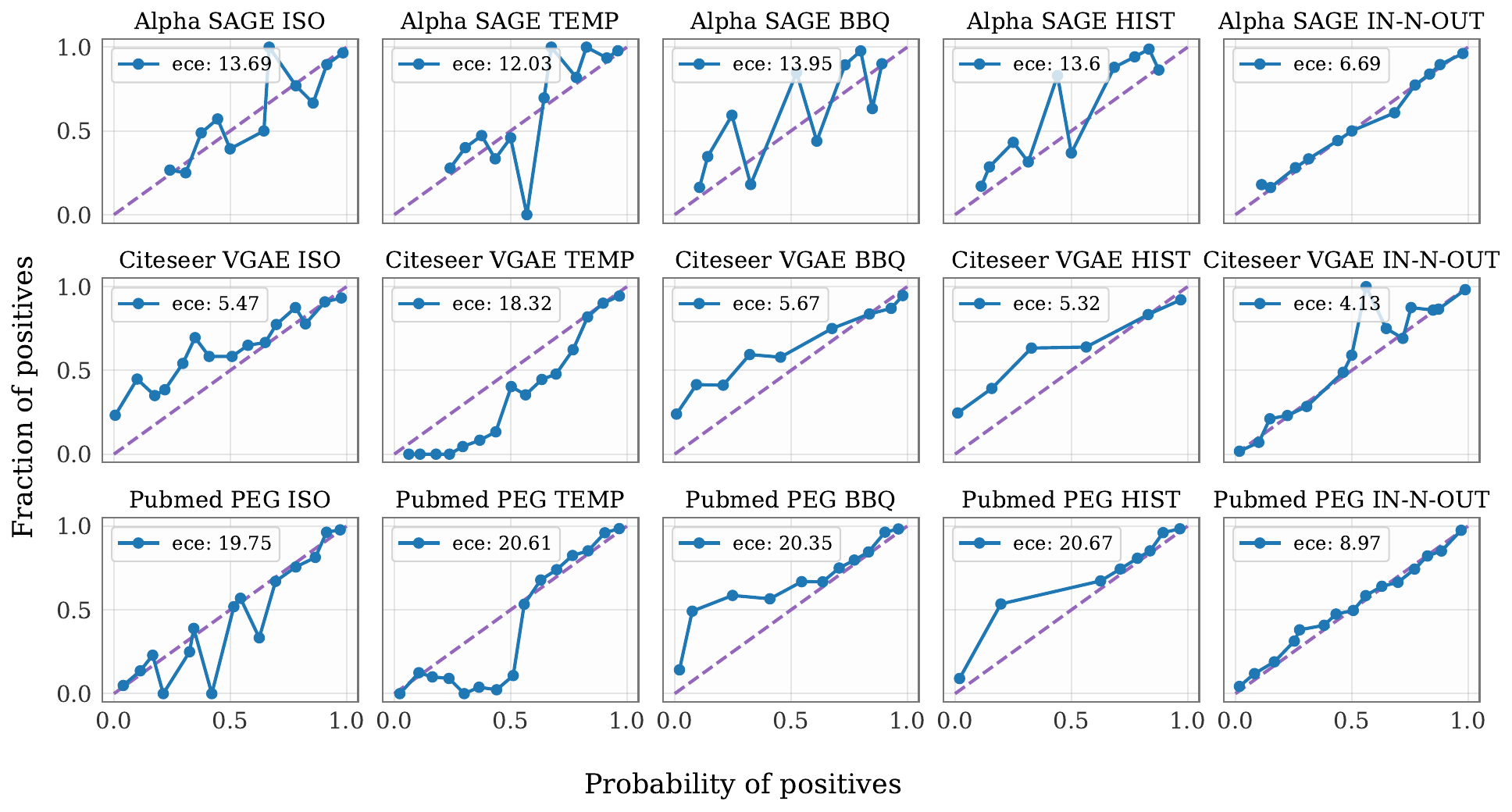}
 \caption{Post-calibration reliability diagrams using \texttt{IN-N-OUT} and off-the-shelf calibration baselines.  \texttt{IN-N-OUT} decreases the ECE of the original GNN models and approaches better the identity lines (denoting a perfectly calibrated model).}
 \label{fig:postCalibration}
\end{figure*}


\vspace{6pt}\noindent\textbf{Baselines.} We compare \texttt{IN-N-OUT} to four calibration methods: Isotonic regression \citep{isotonicRegression}, Temperature scalling \citep{temperatureScalling}, Histogram Binning \citep{histogramBinning}, and Bayesian Binning into Quantiles (BBQ) \citep{BBQ}.
We apply these calibration methods to five GNNs: graph sample and aggregate (GraphSAGE) \citep{SAGE}, variational graph autoencoder \citep[VGAE,][]{VGAE}, positional-encoding GNN \citep[PEG,][]{PEG}, graph convolutional networks \citep[GCN,][]{GCN}, graph isomorphism networks \citep[GIN,][]{xu2018gin}. We chose a diverse set of GNNs to allow for a robust assessment, that generalizes beyond specific architectural choices.  
Rather than benchmarking GNNs on link prediction, we are interested in the calibration aspect of these models. Therefore, we closely follow the original GNNs settings provided in their respective papers. We provide further details in the supplementary material, alongside with the performance of the baseline GNNs on the link prediction datasets.

\subsection{Results and Discussion}

\autoref{tab:ece_results} compares the performance of \texttt{IN-N-OUT} against baselines in terms of ECE. Overall, \texttt{IN-N-OUT} is the best-performing method across all datasets and GNNs. More specifically, our method outperforms the baselines in 29/35 (83\%) of the experiments --- often by a considerable margin. In some scenarios, classic calibration approaches affect negatively the calibration of the baseline (uncalibrated) GNNs. 
This behavior might be associated to these methods assessing structural information only through logits. 
%
%
For a qualitative assessment, \autoref{fig:postCalibration} shows the post-calibration confidence diagrams from ISO, TEMP, BBQ, HIST, and \texttt{IN-N-OUT} on the Citeseer, Pubmed, and Bitcoin-Alpha datasets for the SAGE, VGAE, and PEG models. We observe that, overall, \texttt{IN-N-OUT} better approximates the perfect calibration line compared to the competing methods, which is also reflected in smaller calibration errors (ECE). 
For instance, note that for the SAGE model trained on Bitcoin-Alpha, predictions produced by \texttt{IN-N-OUT} are well-behaved, mostly following the identity line. On the other hand, competing methods produce an erratic behavior with saw-like patterns.

While classic calibration methods do not impact the accuracy of the underlying GNNs, they might affect other metrics commonly used in link prediction. To assess this effect, \autoref{tab:hit20} compares the performance of the methods regarding Hits@20 scores before and after calibration. Due to lack of space in the main manuscript, the supplementary material reports results on additional datasets. Notably, \texttt{IN-N-OUT} has an overall positive effect, sometimes improving Hits@20 scores. On the other hand, using Isotonic Regression, BBQ, and Hist. binning can lead to a significant drop in performance. 

\begin{table*}[thb]
\caption{Hits@20 post-calibration. \textbf{Hist} is the most harmful method for this metric while \texttt{IN-N-OUT} improves it in various scenarios. Here we highlight the worst values in \textcolor{red}{red}.}
\centering
 \resizebox{\textwidth}{!}{ 
\begin{tabular}{l l c c c c c c c c|}
\toprule 
\textbf{Datasets} \color{black} \color{black} & \textbf{Model} \color{black} \color{black} & \textbf{uncalibrated} \color{black} \color{black} & \textbf{Iso} \color{black} \color{black} & \textbf{Temp} \color{black} \color{black} & \textbf{BBQ} \color{black} \color{black} & \textbf{Hist} \color{black} \color{black} & \textbf{IN-N-OUT}\\ 
\midrule
 \parbox[t]{5mm}{\multirow{3}{*}{\rotatebox[origin=c]{0}{Cora}}} \color{black} & SAGE \color{black} &25.08 \color{gray} $\pm$ 1.81 \color{black} &22.46 \color{gray} $\pm$ 3.76 \color{black} &25.01 \color{gray} $\pm$ 1.81 \color{black} &21.12 \color{gray} $\pm$ 2.52 \color{black} &\textcolor{red}{15.42} \color{gray} $\pm$ 1.12 \color{black} &26.89 \color{gray} $\pm$ 1.19\\
 
 \color{black} & PEG \color{black} &67.32 \color{gray} $\pm$ 2.60 \color{black} &\textcolor{red}{51.63} \color{gray} $\pm$ 2.34 \color{black} &67.30 \color{gray} $\pm$ 2.60 \color{black} &54.01 \color{gray} $\pm$ 1.40 \color{black} &51.98 \color{gray} $\pm$ 1.89 \color{black} &68.37 \color{gray} $\pm$ 1.79\\
 
 
 \color{black} & GIN \color{black} &52.18 \color{gray} $\pm$ 1.16 \color{black} &37.00 \color{gray} $\pm$ 1.23 \color{black} &52.18 \color{gray} $\pm$ 1.16 \color{black} &30.17 \color{gray} $\pm$ 1.30 \color{black} &\textcolor{red}{24.28} \color{gray} $\pm$ 1.42 \color{black} &51.23 \color{gray} $\pm$ 1.63\\

 \midrule

 \parbox[t]{5mm}{\multirow{3}{*}{\rotatebox[origin=c]{0}{Citeseer}}}   \color{black} & SAGE \color{black} &38.09 \color{gray} $\pm$ 2.92 \color{black} &36.73 \color{gray} $\pm$ 3.50 \color{black} &38.06 \color{gray} $\pm$ 2.92 \color{black} &33.14 \color{gray} $\pm$ 4.36 \color{black} &\textcolor{red}{27.98} \color{gray} $\pm$ 10.1 \color{black} &40.15 \color{gray} $\pm$ 1.59\\
 
 \color{black} & PEG \color{black} &55.11 \color{gray} $\pm$ 1.59 \color{black} &53.68 \color{gray} $\pm$ 1.74 \color{black} &55.10 \color{gray} $\pm$ 1.59 \color{black} &55.16 \color{gray} $\pm$ 1.76 \color{black} &\textcolor{red}{53.44} \color{gray} $\pm$ 1.56 \color{black} &56.28 \color{gray} $\pm$ 1.61\\
 
 
 \color{black} & GIN \color{black} &38.46 \color{gray} $\pm$ 1.86 \color{black} &38.24 \color{gray} $\pm$ 1.45 \color{black} &38.46 \color{gray} $\pm$ 1.86 \color{black} &\textcolor{red}{32.08} \color{gray} $\pm$ 1.23 \color{black} &33.40 \color{gray} $\pm$ 1.62 \color{black} &38.45 \color{gray} $\pm$ 1.30\\ 

\midrule

 \parbox[t]{7mm}{\multirow{3}{*}{\rotatebox[origin=c]{0}{Alpha}}} \color{black} & SAGE \color{black} &44.97 \color{gray} $\pm$ 2.43 \color{black} &{39.87} \color{gray} $\pm$ 1.34 \color{black} &44.97 \color{gray} $\pm$ 2.43 \color{black} &43.48 \color{gray} $\pm$ 1.43 \color{black} & \textcolor{red}{37.39} \color{gray} $\pm$ 1.13 \color{black} &47.02 \color{gray} $\pm$ 1.09\\

 \color{black} & PEG \color{black} &49.29 \color{gray} $\pm$ 0.89 \color{black} &49.29 \color{gray} $\pm$ 0.91 \color{black} &49.29 \color{gray} $\pm$ 0.89 \color{black} &49.29 \color{gray} $\pm$ 0.96 \color{black} &\textcolor{red}{46.03} \color{gray} $\pm$ 1.01  \color{black} &49.29 \color{gray} $\pm$ 0.79\\
 
 \color{black} & GIN \color{black} &41.50 \color{gray} $\pm$ 1.27 \color{black} &39.73 \color{gray} $\pm$ 1.62 \color{black} &41.50 \color{gray} $\pm$ 1.27 \color{black} &42.91 \color{gray} $\pm$ 1.68 \color{black} &\textcolor{red}{38.17} \color{gray} $\pm$ 1.63 \color{black} &45.25 \color{gray} $\pm$ 1.40\\ 
\bottomrule
\end{tabular}
}
\label{tab:hit20}
\end{table*}

\subsection{Ablation study}
To verify that the strategy of comparing edge embeddings with/without the link of interest is crucial to \texttt{IN-N-OUT}'s performance, \autoref{tab:ablation} compares it against a simple baseline that computes a temperature scale passing `original' edge embeddings through an MLP (Emb.+MLP). Note that, overall, \texttt{IN-N-OUT} consistently outperforms the baseline.

\begin{table}[hbtp!]
\caption{Expected calibration error (mean and standard deviation). The best ECE results found for each GNN are highlighted in \textcolor{blue}{blue}. In most cases, \texttt{IN-N-OUT} outperforms the baseline by a considerable margin.}
\centering
\begin{tabular}{l l r r r}
\toprule
\color{black} & \textbf{Model} \color{black} & \textbf{Uncalibrated} \color{black} & \textbf{Emb.+MLP} \color{black} & \textbf{{IN-N-OUT}} \\ 
\midrule
 \parbox[t]{5mm}{\multirow{3}{*}{\rotatebox[origin=c]{0}{Cite}}}
 \color{black} & VGAE \color{black} & 14.00 \color{gray} $\pm$ 0.67 \color{black} & 4.92 \color{gray} $\pm$ 1.06 \color{black} & \textbf{\color{blue}4.19} \color{gray} $\pm$ 0.65 \\
 \color{black} & SAGE \color{black} & 14.52 \color{gray} $\pm$ 1.89 \color{black} & \textbf{\color{blue}2.62} \color{gray} $\pm$ 0.62 \color{black} & 3.79 \color{gray} $\pm$ 1.98 \\
 \color{black} & PEG \color{black} & 28.39 \color{gray} $\pm$ 1.59 \color{black} & 17.75 \color{gray} $\pm$ 1.21 \color{black} & \textbf{\color{blue}12.02} \color{gray} $\pm$ 0.98 \\
\midrule
 \parbox[t]{5mm}{\multirow{3}{*}{\rotatebox[origin=c]{0}{Pub}}}
 \color{black} & VGAE \color{black} & 20.41 \color{gray} $\pm$ 0.64 \color{black} & 3.01 \color{gray} $\pm$ 1.16 \color{black} & \textbf{\color{blue}1.85} \color{gray} $\pm$ 0.37  \\
 \color{black} & SAGE \color{black} & 18.78 \color{gray} $\pm$ 1.42 \color{black} & 4.16 \color{gray} $\pm$ 1.47 \color{black} &  \textbf{\color{blue}3.01} \color{gray} $\pm$ 1.32 \\
\color{black} & PEG \color{black} & 20.49 \color{gray} $\pm$ 1.31 \color{black} & 8.90 \color{gray} $\pm$ 0.89 \color{black} & \textbf{\color{blue}8.21} \color{gray} $\pm$ 0.75 \\
\midrule
 \parbox[t]{5mm}{\multirow{3}{*}{\rotatebox[origin=c]{0}{TW}}}
 \color{black} & VGAE \color{black} & 16.32 \color{gray} $\pm$ 0.20 \color{black} & 4.81 \color{gray} $\pm$ 1.56 \color{black} & \textbf{\color{blue}1.50} \color{gray} $\pm$ 0.44 \\
 \color{black} & SAGE \color{black} & 16.84 \color{gray} $\pm$ 2.84 \color{black} & 4.06 \color{gray} $\pm$ 1.75 \color{black} & \textbf{\color{blue}1.53} \color{gray} $\pm$ 0.89 \\
 \color{black} & PEG \color{black} & 7.90 \color{gray} $\pm$ 0.91 \color{black} & 2.95 \color{gray} $\pm$ 0.59 \color{black} & \textbf{\color{blue}2.21} \color{gray} $\pm$ 0.51 \\
\midrule
 \parbox[t]{7mm}{\multirow{3}{*}{\rotatebox[origin=c]{0}{Chame}}}
 \color{black} & VGAE \color{black} & 7.52 \color{gray} $\pm$ 1.63 \color{black} & 3.57 \color{gray} $\pm$ 1.19 \color{black} & \textbf{\color{blue}1.07} \color{gray} $\pm$ 0.34 \\
 \color{black} & SAGE \color{black} & 5.98 \color{gray} $\pm$ 3.61 \color{black} & 3.99 \color{gray} $\pm$ 1.45 \color{black} & \textbf{\color{blue}2.46} \color{gray} $\pm$ 0.86 \\
 \color{black} & PEG \color{black} & 4.44 \color{gray} $\pm$ 0.24 \color{black} & 2.14 \color{gray} $\pm$ 0.21 \color{black} & \textbf{\color{blue}1.66} \color{gray} $\pm$ 0.36 \\
\midrule
  \parbox[t]{7mm}{\multirow{3}{*}{\rotatebox[origin=c]{0}{Alpha}}}
  \color{black} & VGAE \color{black} & 20.36 \color{gray} $\pm$ 1.53 \color{black} &  \textbf{\color{blue}7.12} $\pm$ \color{gray} 1.19  \color{black} & 8.42 $\pm$ \color{gray} 0.68\\
  \color{black} & SAGE \color{black} & 21.26 $\pm$ \color{gray} 1.37 \color{black} & 7.82 $\pm$ \color{gray}  1.45 \color{black} & \textbf{\color{blue}6.68} $\pm$ \color{gray} 0.91\\
  \color{black} & PEG \color{black} & 13.91 $\pm$ \color{gray} 1.13 \color{black} & \textbf{\color{blue}7.62} $\pm$ \color{gray} 1.76 \color{black} & 9.78 $\pm$ \color{gray} 0.89\\
\bottomrule
\end{tabular}
\label{tab:ablation}
\end{table}
\section{Conclusion} 
This is the first work on the calibration of GNNs in link prediction tasks. First, we empirically analyzed the confidence prediction patterns for several combinations of GNNs and datasets. We observed that GNNs are often miscalibrated and may depict complex confidence patterns that escape the usual patterns of consistent over- or underconfidence. In particular, the calibration patterns mainly show a dependency on the predicted labels: class-0 predictions tend to be underconfident, whereas class-1 ones are usually overconfident. Based on this insight, we proposed an elegant yet effective temperature-scaling approach for GNNs, called \texttt{IN-N-OUT}. Our method connects this insight with the idea that embeddings carry all the information a model has regarding node relationships --- therefore, seeing edges that wildly contradict our predictions should impact embeddings significantly. 

Experiments show that \texttt{IN-N-OUT} outperforms classic calibration methods in a combination of 7 datasets and 5 GNNs. By better quantifying uncertainty in GNNs, we believe this works represents an important step towards reliable graph ML methods, with a potential broad impact across diverse application domains.

While our work focused on link prediction, an interesting direction for future work consists of assessing calibration in other tasks like graph classification, temporal link prediction, and dynamic node classification.

\section*{Acknowledgements}
Erik Nascimento was partially supported by the Fundação Cearense de Apoio ao Desenvolvimento Científico e Tecnológico (FUNCAP).
Diego Mesquita acknowledges the support by the Silicon Valley Community Foundation (SVCF) through the Ripple impact fund,  the Funda\c{c}\~ao de Amparo \`a Pesquisa do Estado do Rio de Janeiro (FAPERJ) through the \emph{Jovem Cientista do Nosso Estado} program, the Funda\c{c}\~ao de Amparo \`a Pesquisa do Estado de São Paulo (FAPESP) grant 2023/00815-6, and the Conselho Nacional de Desenvolvimento Científico e Tecnológico (CNPq) grant 404336/2023-0.  
Amauri H. Souza and Samuel Kaski were supported by the Academy of Finland (Flagship programme: Finnish Center for Artificial Intelligence FCAI), EU Horizon 2020 (European Network of AI Excellence Centres ELISE, grant agreement 951847), UKRI Turing AI World-Leading Researcher Fellowship (EP/W002973/1).
We also acknowledge the computational resources provided by the Aalto Science-IT Project from Computer Science IT.


\bibliographystyle{IEEEtranN}
\bibliography{references}

\appendix

\section{Datasets and Implementation details}

\autoref{tab:details_datasets} shows summary statistics for the datasets used in our experiments.
We chose the optimal hyperparameters for all GNNs in this work (see \autoref{tab:best_hyper}) using grid search. 
We consider learning rate values in $\{10^{-1}, 10^{-2}, 10^{-3}\}$, number of epochs in $\{400, 600, 800, 1000\}$ and up to three convolutional layers outputing embeddings of size in $\{16, 32, 64, 128\}$. The embedding size of for the last convolutional layer (out) is chosen independently from the previous layers, which have same dimension (hidden). GCN, GIN, and VGAE use inner products between node embeddings to compute logits for link prediction. SAGE and PEG use a point-wise product of node embeddings followed by a linear layer.
We used an 80/10/10\% (train/val/test) split for all datasets. For the confidence calibration phase, we train \texttt{IN-N-OUT} using Adam with learning rate $1 \times 10^{-4}$ with a weight decay $5.0 \times 10^{-8}$ for $5000$ epochs. Similarly to \citet{CaGCN} and \citet{GATS}, we trained all calibrators using the training set partition. We train each GNN (plus dataset) 5 times, and for each time, we calibrate the resulting model 5 times. This adds up to 25 rounds of experiments per dataset/GNN combination. The calibration baselines were implemented with the Netcal package \citep{netcal}. \texttt{IN-N-OUT} and all GNNs were implemented in Pytorch. We have run all experiments in a computer equipped with a consumar-grade GPU (Nvidia RTX 3060 6GB), an Intel Core i7-12700H 12th generation CPU, and 16GB DDR5 RAM.
%

\begin{table}[H]
\centering
\caption{Summary statistics of the datasets.}
\begin{tabular}{lcccc}
\toprule
\textbf{Dataset} & \textbf{\# Nodes} & \textbf{\# Edges} & \textbf{\# Feat.} & \textbf{\# Labels}\\
\midrule
Cora & 2708 & 10556 & 1433 & 7 \\
PubMed & 19717 & 88648 & 500 & 3\\
Citeseer & 3327 & 9104 & 3703 & 6\\
Twitch & 1912 & 57905 & 128 & 2\\
Chameleon & 2277 & 39730 & 2325 & 5\\
Bitcoin-Alpha & 3783 & 28248 & 8 & 2\\
Bitcoin-OTC & 5881 & 42984 & 8 & 2\\
OGBL-DDI & 4267 & 1334889 & 5 & 2\\ 	
OGBL-Collab & 235868 & 1285465 & 128 & 2\\ 		
\bottomrule
\end{tabular}
\label{tab:details_datasets}
\end{table}

\begin{table}[!ht]
\caption{\small Optimal hyper-parameters for the GNNs. }
\centering
\begin{tabular}{l  c c c c c c c}
\toprule
& \textbf{Model} & \textbf{hidden}& \textbf{out}& \textbf{layers}& \textbf{lr}&  \textbf{epochs}\\ 
\cmidrule{1-7}
  & GIN  & $64$ & $16$ & $2$ & $0.01$ &  $1000$\\
 & SAGE & $128$ & $64$ & $2$ & $0.01$ &  $1000$\\
 & VGAE  & $32$ & $16$ & $2$ & $0.001$ &  $400$\\
 & PEG &  $20$ & $16$ & $2$ & $0.01$ &  $400$\\
 & GCN  & $32$ & $16$ & $2$ & $0.001$ &  $400$\\
 & SEAL  & $32$ & $32$ & $3$ & $0.0001$ &  $50$\\
\bottomrule
\end{tabular}
\label{tab:best_hyper}
\vspace{-0.35cm}
\end{table}

\section{Time and space complexity of \texttt{IN-N-OUT}}

Once \texttt{IN-N-OUT} is trained, its forward pass involves the computation of auxiliary edge embeddings with/without counting edge of interest, as well as a subsequent forward pass of $\mathrm{MLP}_{c_1}$ or $\mathrm{MLP}_{c_2}$. That being said, since $\mathrm{MLP}_{c_1}$/$\mathrm{MLP}_{c_2}$'s are constants in the size of the input graph, $\texttt{IN-N-OUT}$ does not increase the asymptotic time/space complexity of the baseline GNNs.

\section{Additional results}

For each of the 5 GNNs, we run post-hoc calibration 5 times. To confirm the statistical significance of our results, we have run Wilcoxon's signed rank test on the difference between In-N-Out and the second-best calibrator (\autoref{tab:hipotese}) --- for cases when In-N-Out is the best calibrator (non-empty entries of Table C). In approximately $80\%$ of the cases, {\textcolor{black}{results are statistically significant}} (denoted by \textcolor{blue}{\checkmark}).

\begin{table}[hbtp!]
\caption{Wilcoxon’s sign rank tests \(\alpha = 5\%\).} 
\centering

\begin{tabular}{lccccccc}
\toprule
\textbf{} & \textbf{cora} & \textbf{cite}  & \textbf{pub} & \textbf{twitch} & \textbf{chame} & \textbf{alpha} & \textbf{otc}\\ 
\toprule
VGAE & {\color{blue} \checkmark} & \xmark & {\color{blue} \checkmark} & {\color{blue} \checkmark} & {\color{blue} \checkmark} & \xmark &  \\
SAGE & {\color{blue} \checkmark} & {\color{blue} \checkmark} & {\color{blue} \checkmark} & \xmark & \xmark & {\color{blue} \checkmark} &  \\
PEG & {\color{blue} \checkmark} & {\color{blue} \checkmark} & {\color{blue} \checkmark} & {\color{blue} \checkmark} & {\color{blue} \checkmark} & {\color{blue} \checkmark} & {\color{blue} \checkmark} \\
GIN & {\color{blue} \checkmark} & {\color{blue} \checkmark} & {\color{blue} \checkmark} &  & {\color{blue} \checkmark} & {\color{blue} \checkmark} & {\color{blue} \checkmark} \\
GCN & {\color{blue} \checkmark} & {\color{blue} \checkmark} &  &  &  & \xmark & \xmark \\
\bottomrule
\end{tabular}
\label{tab:hipotese}
\end{table}
We report the predictive performance of the GNNs before calibration. \autoref{tab:performanceGNN} shows AUC ROC (Area Under the Receiver Operator Characteristic curve), Hits@20, and accuracy values for all GNNs and datasets. We note that the best-performing method varies with the metric. For instance, on Cora, GCN is the winner if we consider AUC whereas GIN achieves the highest accuracy. On three out of seven datasets, GraphSAGE achieves the lowest average AUC.

\begin{figure*}[h]
\centering
\includegraphics[width=\linewidth]{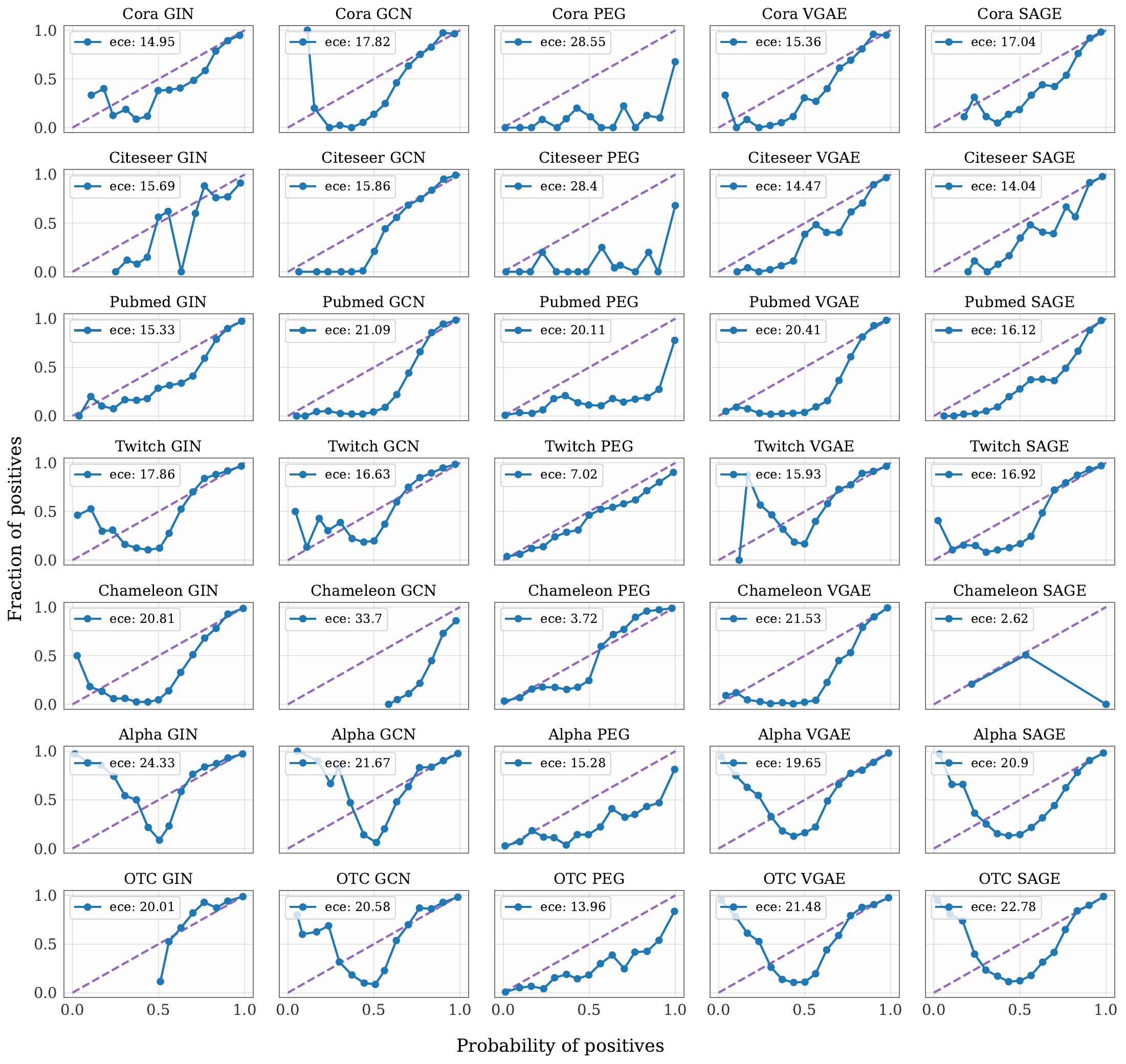}
 \caption{Pre-calibration reliability diagram for all models and datasets.}
 \label{fig:pre-calibration-all}
\end{figure*}

\begin{figure*}[h]
\centering
\includegraphics[width=\linewidth]{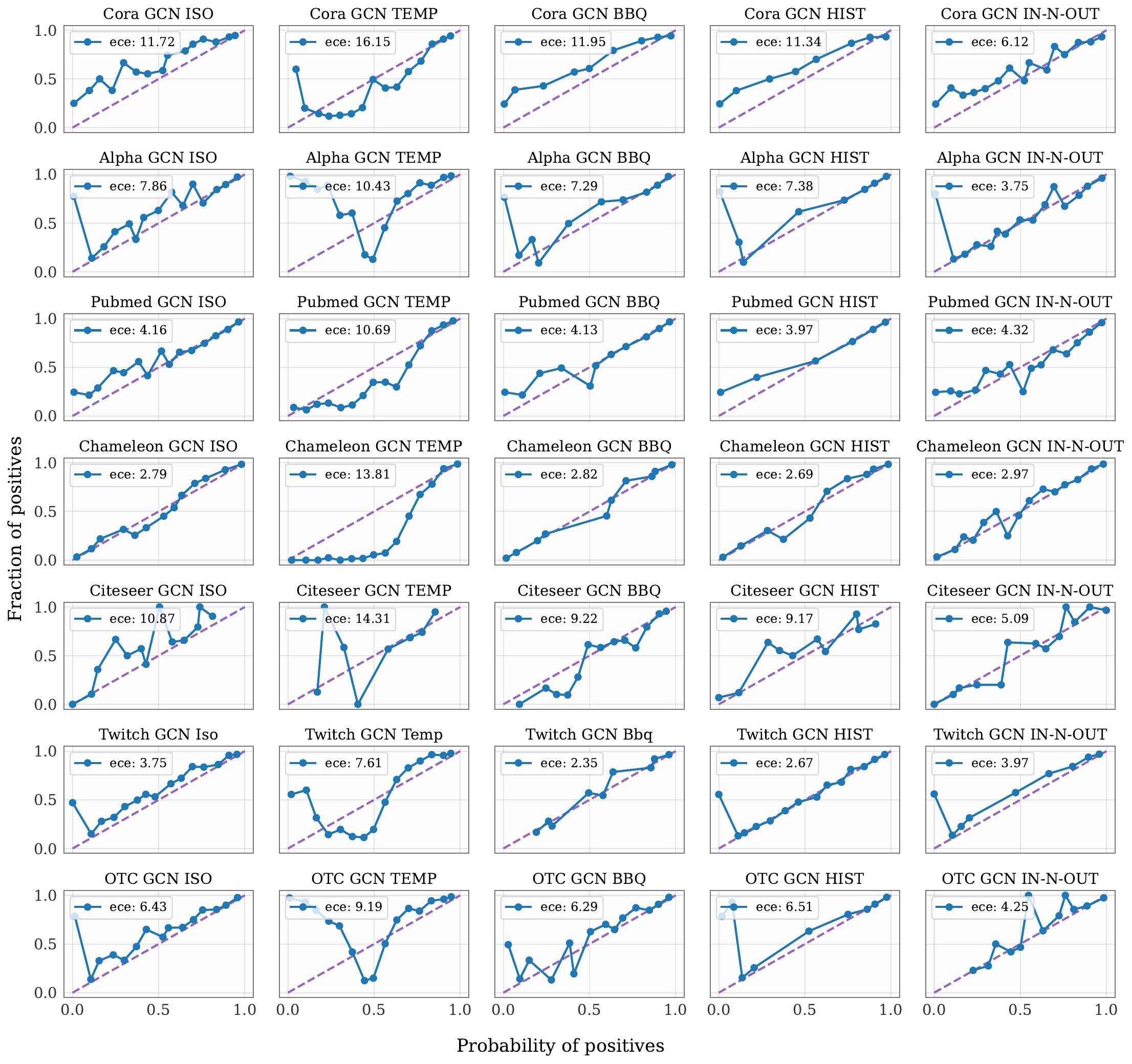}
 \caption{Post calibration reliability diagram for GCN and all datasets. Overall, IN-N-OUT reduced the ECE of all GNNs, resulting in more well-behaved reliability diagrams when compared to \autoref{fig:pre-calibration-all}.}
 \label{fig:post-calibration-all}
\end{figure*}

\begin{table*}[!th]
\caption{\small GNN performances prior to calibration (mean and standard deviation for AUC, Hits@20 and ACC).}
\centering
\begin{tabular}{l l c c c c|}
\toprule
\textbf{Dataset} & \textbf{Model} & \textbf{AUC} & \textbf{Hits@20}& \textbf{ACC}\\ 
\cmidrule{1-5}

 \parbox[t]{7mm}{\multirow{6}{*}{\rotatebox[origin=c]{0}{cora}}}
 & VGAE & $89.79 \pm 0.50$ & $ 54.70\pm 4.40$ & $74.57 \pm 0.76$\\
 & SAGE & $87.93 \pm 1.68$ & $61.48 \pm 1.81$ & $75.23 \pm 1.74$\\
 & PEG & $88.01 \pm 1.76$ & $61.85 \pm 1.38$ & $71.06 \pm 1.25$ \\
 & GIN & $89.17 \pm 1.54$ & $52.18 \pm 1.16$ & $80.64 \pm 1.67$\\
 & GCN & $89.23 \pm 1.79$ & $62.80 \pm 1.72$ & $78.46 \pm 1.85$\\
 & SEAL & $91.24 \pm 0.89$ & $68.29 \pm 0.75$ & $79.68 \pm 0.83$\\
 
\midrule

 \parbox[t]{7mm}{\multirow{6}{*}{\rotatebox[origin=c]{0}{cite}}}
 & VGAE & $86.36 \pm 2.37$ & $50.95 \pm 5.21$ & $74.61 \pm 0.67$\\
 & SAGE & $84.89 \pm 1.47$ & $56.92 \pm 2.92$ & $72.19 \pm 1.89$\\
 & PEG & $90.57 \pm 1.62$ & $61.53 \pm 2.13$ & $70.65 \pm 1.49$\\
 & GIN & $82.47 \pm 1.76$ & $38.46 \pm 1.86$ & $78.90 \pm 1.81$\\
 & GCN & $82.24 \pm 1.91$ & $58.68 \pm 1.10$ & $76.37 \pm 1.67$\\
 & SEAL & $88.06 \pm 0.72$ & $67.28 \pm 0.91$ & $78.31 \pm 0.92$\\
 
\midrule

 \parbox[t]{7mm}{\multirow{6}{*}{\rotatebox[origin=c]{0}{pub}}}
 & VGAE & $95.37 \pm 0.37$ & $37.87 \pm 2.92$ & $75.49 \pm 0.64$\\
 & SAGE & $89.15 \pm 2.28$ & $41.85 \pm 1.99$ & $73.68 \pm 1.42$\\
 & PEG & $93.74 \pm 2.76$ & $44.35 \pm 1.13$ & $77.96 \pm 1.34$\\
 & GIN & $82.50 \pm 1.79$ & $26.96 \pm 1.24$ & $73.69 \pm 1.45$\\
 & GCN & $84.54 \pm 2.16$ & $37.65 \pm 2.12$ & $78.65 \pm 2.10$\\
 & SEAL & $96.81 \pm 0.26$ & $52.67 \pm 0.39$ & $82.97 \pm 0.34$\\
 
\midrule

 \parbox[t]{7mm}{\multirow{5}{*}{\rotatebox[origin=c]{0}{Twitch}}}
 & VGAE & $83.70 \pm 0.26$ & $19.47 \pm 0.24$ & $16.3 \pm 0.20$\\
 & SAGE & $88.03 \pm 0.68$ & $20.70 \pm 1.41$ & $69.34 \pm 0.84$\\
 & PEG & $91.68 \pm 0.72$ & $22.94 \pm 1.79$ & $84.69 \pm 0.91$\\
 & GIN & $87.64 \pm 1.21$ & $20.96 \pm 1.73$ & $65.85 \pm 1.12$\\
 & GCN & $85.49 \pm 1.79$ & $24.89 \pm 2.03$ & $66.53 \pm 2.16$\\

\midrule

 \parbox[t]{7mm}{\multirow{6}{*}{\rotatebox[origin=c]{0}{chame}}}
 & VGAE & $97.22 \pm 1.17$ & $62.04 \pm 2.81$ & $58.70 \pm 1.63$\\
 & SAGE & $53.62 \pm 3.31$ & $23.8 \pm 0.88$ & $51.62 \pm 3.12$\\
 & PEG & $91.28 \pm 0.98$ & $46.67 \pm 0.82$ & $75.81 \pm 0.97$\\
  & GIN & $97.06 \pm 0.94$ & $71.45 \pm 1.08$ & $79.96 \pm 0.87$\\
 & GCN & $87.96 \pm 0.76$ & $13.26 \pm 1.96$ & $72.16 \pm 0.79$\\
 & SEAL & $98.12 \pm 0.54$ & $74.26 \pm 0.64$ & $82.13 \pm 0.44$\\
 
\midrule

 \parbox[t]{7mm}{\multirow{6}{*}{\rotatebox[origin=c]{0}{alpha}}}
 & VGAE & $81.91 \pm 2.21$ & $44.05 \pm 2.92$ & $64.37 \pm 2.26$\\
 & SAGE & $75.44 \pm 2.14$ & $37.46 \pm 1.23$ & $64.02 \pm 1.59$\\
  & PEG & $90.49 \pm 1.09$ & $47.87 \pm 1.19$ & $82.43 \pm 1.07$\\
 & GIN & $79.15 \pm 1.28$ & $41.50 \pm 1.27$ & $58.64 \pm 1.38$\\
 & GCN & $86.96 \pm 1.81$ & $45.32 \pm 1.97$ & $57.60 \pm 2.61$\\
 & SEAL & $93.91 \pm 0.41$ & $52.97 \pm 0.38$ & $87.28 \pm 0.54$\\

 \midrule

 \parbox[t]{7mm}{\multirow{6}{*}{\rotatebox[origin=c]{0}{OTC}}}
 & VGAE & $82.63\pm 2.76$ & $45.13 \pm 2.36$ & $64.02 \pm 2.53$\\
 & SAGE & $79.80 \pm 0.87$ & $44.48 \pm 0.67$ & $65.65 \pm 0.95$\\
 & PEG & $91.94 \pm 1.16$ & $59.56 \pm 1.05$ & $83.80 \pm 0.76$\\
 & GIN & $93.00 \pm 1.91$ & $46.39 \pm 1.98$ & $50.27 \pm 1.93$\\
 & GCN & $88.41 \pm 0.94$ & $44.62 \pm 1.87$ & $57.60 \pm 0.97$\\
 & SEAL & $95.94 \pm 0.61$ & $64.63 \pm 0.44$ & $84.34 \pm 0.62$\\

 \midrule

 \parbox[t]{7mm}{\multirow{2}{*}{\rotatebox[origin=c]{0}{DDI}}}
 & VGAE & $76.53 \pm 0.28$ & $37.12 \pm 0.14$ & $55.27 \pm 0.31$\\
 & PEG & $88.17 \pm 0.49$ & $39.07 \pm 0.20$ & $62.91 \pm 0.49$\\

  \midrule

 \parbox[t]{7mm}{\multirow{2}{*}{\rotatebox[origin=c]{0}{Collab}}}
 & VGAE & $72.26 \pm 0.46$ & $31.14 \pm 0.26$ & $46.81 \pm 0.44$\\
 & PEG & $85.57 \pm 0.42$ & $34.22 \pm 0.30$ & $48.09 \pm 0.39$\\
 
\bottomrule
\end{tabular}
\label{tab:performanceGNN}
\vspace{-0.35cm}
\end{table*}
\autoref{fig:pre-calibration-all} shows the reliability diagrams (before calibration) for all GNNs and datasets. Following the trend discussed in the main paper, most diagrams depict a prediction-dependent behavior (overconfident for the positive class and underconfident for the negative one). The second most common pattern represents overconfident predictions for both negative and positive predictions.
Moreover, \autoref{fig:post-calibration-all} reports post-calibration reliability diagrams. Here, \texttt{IN-N-OUTS} outperforms the baselines with respect to ECE by a large margin in 3 out 7 datasets.

\autoref{tab:hit20app} shows Hits@20 values for all models after calibration. Overall, histogram binning achieves the lowest (worst) values, sometimes reaching zero.  
In such cases, we observe that GNNs assign a probability greater than 0.9 to the twentieth largest negative edge (VGAE does this for Pubmed or Twitch, for example), and these values are within the $(0.9, 1.0]$ bin. 
%
{As histogram binning collapses the confidence values in a bin to the average of the values in that bin, then Hits@20 will be zero in these cases. Note that this is a behavior that depends on both the GNN and the calibration method.}

\begin{table*}[!ht]
\caption{Hits@20 post-calibration. The lowest values are highlighted in red. \textbf{BBQ} and \textbf{Hits} are the most harmful methods for this metric. Importantly, \texttt{IN-N-OUT} improves both metrics in most cases ($>70$\%). Blue denotes cases in which \texttt{IN-N-OUT} improves Hits@20.}
\centering
\resizebox{\textwidth}{!}{
\begin{tabular}{l l c c c c c c c c|}
\toprule
& \textbf{Model} & \textbf{uncalibrated} & \textbf{Iso} & \textbf{Temp} & \textbf{BBQ} & \textbf{Hist} & \textbf{IN-N-OUT}\\ 

\midrule

 \parbox[t]{5mm}{\multirow{6}{*}{\rotatebox[origin=c]{0}{cora}}}
 & VGAE & $54.70\pm 4.40$ & $50.89 \pm 4.27$ & $54.70 \pm 4.40$ & $49.34 \pm 4.67$ & \color{red}{$\bm{45.12 \pm 3.94}$} & \color{blue}{$55.62 \pm 3.15$}\\
 
 & SAGE & $61.48 \pm 1.81$ & $58.12 \pm 3.76$ & $61.48 \pm 1.81$ & $58.06 \pm 3.71$ & \color{red}{$\bm{49.93 \pm 2.76}$} & $61.17 \pm 2.72$\\
 
 & PEG & $61.85 \pm 1.38$ & $57.21 \pm 2.97$ & $61.85 \pm 1.38$ & $54.78 \pm 2.3$ & \color{red}{$\bm{52.98 \pm 1.89}$} & \color{blue}{$62.39 \pm 2.57$}\\
 
 & GIN & $52.18 \pm 1.16$ & $37.00 \pm 1.23$ & $52.18 \pm 1.16$ & $30.17 \pm 1.30$ & \color{red}{$\bm{24.28 \pm 1.42}$} & $51.23 \pm 1.63$\\

 & GCN & $62.80 \pm 1.72$ & $60.15 \pm 1.92$ & $62.80 \pm 1.72$ & $62.61 \pm 1.69$ & \color{red}{$\bm{57.68 \pm 1.39}$} & $62.42 \pm 1.21$\\

 & SEAL & $71.91 \pm 0.92$ & $70.20 \pm 0.98$ & $71.91 \pm 0.92$ & $71.53 \pm 0.85$ & \color{red}{$\bm{68.69 \pm 0.89}$} & $71.98 \pm 0.80$\\
 
\midrule

 \parbox[t]{5mm}{\multirow{6}{*}{\rotatebox[origin=c]{0}{cite}}}
 & VGAE & $50.95 \pm 5.21$ & $47.98 \pm 4.36$ & $50.95 \pm 5.21$ & $49.79 \pm 4.39$ & \color{red}{$\bm{43.18 \pm 4.87}$}  & \color{blue}{$52.15 \pm 3.97$}\\
 
 & SAGE & $56.92 \pm 2.92$ & $53.97 \pm 2.98$ & $56.92 \pm 2.92$ & $49.53 \pm 4.21$ & \color{red}{$\bm{35.75 \pm 4.13}$} & $56.89 \pm 2.13$\\
 
 & PEG & $61.53 \pm 2.13$ & $56.11 \pm 3.27$ & $61.53 \pm 2.13$ & $53.23 \pm 2.61$ & \color{red}{$\bm{51.14 \pm 2.48}$} & \color{blue}{$62.20 \pm 2.09$}\\
 
 & GIN & $38.46 \pm 1.86$ & $38.24 \pm 1.45$ & $38.46 \pm 1.86$ & \color{red}{$\bm{32.08 \pm 1.23}$} & $33.40 \pm 1.62$ & \color{blue}{$38.47 \pm 1.30$}\\

 & GCN & $58.68 \pm 1.10$ & $58.24 \pm 1.60$ & $58.68 \pm 1.10$ & $57.14 \pm 1.25$ & \color{red}{$\bm{47.47 \pm 1.62}$} & \color{blue}{$60.21 \pm 1.58$}\\

 & SEAL & $72.48 \pm 0.68$ & $72.48 \pm 0.71$ & $72.48 \pm 0.68$ & $71.72 \pm 0.93$ & \color{red}{$\bm{70.96 \pm 0.67}$} & $72.48 \pm 0.63$\\
 
\midrule

 \parbox[t]{5mm}{\multirow{6}{*}{\rotatebox[origin=c]{0}{pub}}}
 & VGAE & $37.87 \pm 2.92$ & $37.66 \pm 2.54$ & $37.87 \pm 2.92$ & $35.76 \pm 2.27$ & \color{red}{$\bm{0.00 \pm 0.00}$}  & \color{blue}{$38.12 \pm 2.25$}\\
 
 & SAGE & $41.85 \pm 1.99$ & $39.79 \pm 3.41$ & $41.85 \pm 1.99$ & $25.76 \pm 3.70$ & \color{red}{$\bm{23.86 \pm 4.10}$} & \color{blue}{$41.97 \pm 2.34$}\\
 
 & PEG & $44.35 \pm 1.13$ & $42.76 \pm 3.89$ & $44.35 \pm 1.13$ & $41.09 \pm 2.73$ & \color{red}{$\bm{0.00 \pm 0.00}$} & \color{blue}{$45.35 \pm 3.02$}\\
 
 & GIN & $26.96 \pm 1.24$ & $23.87 \pm 1.62$ & $26.96 \pm 1.24$ & $0.00 \pm 0.00$ & \color{red}{$\bm{0.00 \pm 0.00}$} & $26.92 \pm 1.63$\\

 & GCN & $37.65 \pm 2.12$ & $34.83 \pm 1.64$ & $37.65 \pm 2.12$ & $28.97 \pm 1.96$ & \color{red}{$\bm{0.00 \pm 0.00}$} & \color{blue}{$38.19 \pm 1.35$}\\

 & SEAL & $72.48 \pm 0.64$ & $72.48 \pm 0.68$ & $72.48 \pm 0.64$ & $70.96 \pm 0.63$ & \color{red}{$\bm{70.96 \pm 0.77}$} & $72.48 \pm 0.62$\\
 
\midrule

 \parbox[t]{5mm}{\multirow{5}{*}{\rotatebox[origin=c]{0}{Twitch}}}
 & VGAE & $19.47 \pm 0.24$ & $13.50 \pm 0.21$ & $19.42 \pm 0.24$ & $0.04 \pm 0.00$ & \color{red}{$\bm{0.04 \pm 0.00}$} & \color{blue}{$19.66 \pm 1.12$}\\
 
 & SAGE & $20.70 \pm 1.41$ & $9.49 \pm 2.18$ & $20.70 \pm 1.41$ & $0.00 \pm 0.00$ & \color{red}{$\bm{0.00 \pm 0.00}$} & \color{blue}{$20.28 \pm 1.14$}\\
 
 & PEG & $22.94 \pm 1.79$ & $21.72 \pm 1.34$ & $22.94 \pm 1.79$ & $0.00 \pm 0.00$ & \color{red}{$\bm{0.00 \pm 0.00}$} & $22.21 \pm 2.32$\\
 
 & GIN & $20.96 \pm 1.73$ & $18.66 \pm 1.53$ & $20.96 \pm 1.73$ & $20.54 \pm 1.42$ & \color{red}{$\bm{0.00 \pm 0.00}$} & \color{blue}{$21.09 \pm 1.50$}\\

 & GCN & $24.89 \pm 2.03$ & \color{red}{$\bm{23.45 \pm 1.43}$} & $24.89 \pm 2.03$ & $23.64 \pm 1.19$ & $24.16 \pm 1.26$ & $24.80 \pm 1.09$\\

\midrule

 \parbox[t]{7mm}{\multirow{6}{*}{\rotatebox[origin=c]{0}{chame}}}
 & VGAE & $62.04 \pm 2.81$ & $52.73 \pm 8.91$ & $62.02 \pm 2.81$ & $0.00 \pm 0.00$ & \color{red}{$\bm{0.00 \pm 0.00}$} & \color{blue}{$64.09 \pm 1.01$}\\
 
 & SAGE & $23.83 \pm 0.88$ & \color{red}{$\bm{19.14 \pm 2.40}$} & $23.80 \pm 0.88$ & $22.07 \pm 2.63$ & $22.26 \pm 3.05$ & $23.63 \pm 1.20$\\

 & PEG & $46.67 \pm 0.82$ & $45.96 \pm 1.58$ & $46.67 \pm 0.82$ & \color{red}{$\bm{42.73 \pm 2.03}$} & $42.87 \pm 1.90$  & \color{blue}{$47.23 \pm 2.34$}\\
 
 & GIN & $71.45 \pm 1.08$ & $68.33 \pm 1.34$ & $71.45 \pm 1.08$ & $71.24 \pm 1.97$ & \color{red}{$\bm{65.63 \pm 1.72}$} & \color{blue}{$72.56 \pm 1.25$}\\

 & GCN & $13.26 \pm 1.96$ & $11.03 \pm 1.57$ & $13.26 \pm 1.96$ & $0.00 \pm 0.00$ & \color{red}{$\bm{0.00 \pm 0.00}$} & $13.25 \pm 1.09$\\

 & SEAL & $74.26 \pm 0.64$ & $74.23 \pm 0.49$ & $74.26 \pm 0.64$ & $70.13 \pm 0.12$ & \color{red}{$\bm{69.78 \pm 0.19}$} & $74.26 \pm 0.53$\\
 
\midrule

 \parbox[t]{7mm}{\multirow{6}{*}{\rotatebox[origin=c]{0}{Alpha}}}
 & VGAE & $44.05 \pm 2.92$ & $44.01 \pm 3.41$ & $44.05 \pm 2.92$ & $40.91 \pm 2.72$ & \color{red}{$\bm{0.00 \pm 0.00}$} & \color{blue}{$44.96 \pm 2.19$}\\
 
 & SAGE & $37.46 \pm 1.23$ & $35.18 \pm 2.16$ & $37.46 \pm 1.23$ & $34.76 \pm 2.63$ & \color{red}{$\bm{21.34 \pm 2.34}$} & \color{blue}{$38.06 \pm 1.67$}\\

 & PEG & $47.87 \pm 1.19$ & $43.51 \pm 2.52$ & $47.87 \pm 1.19$ & $44.36 \pm 2.14$ & \color{red}{$\bm{34.72 \pm 2.63}$}  & \color{blue}{$47.89 \pm 1.86$}\\
 
 & GIN & $41.50 \pm 1.27$ & $39.73 \pm 1.62$ & $41.50 \pm 1.27$ & $42.91 \pm 1.68$ & \color{red}{$\bm{38.17 \pm 1.63}$} & \color{blue}{$45.25 \pm 1.40$}\\

 & GCN & $45.32 \pm 1.97$ & $44.33 \pm 1.92$ & $45.32 \pm 1.97$ & $44.61 \pm 1.92$ & \color{red}{$\bm{39.02 \pm 1.87}$} & \color{blue}{$45.49 \pm 1.93$}\\

 & SEAL & $52.97 \pm 0.66$ & $50.63 \pm 0.45$ & $52.97 \pm 0.66$ & \color{red}{$\bm{47.73 \pm 0.61}$} & $52.19 \pm 0.33$ & $52.97 \pm 0.5$\\
 
\midrule

 \parbox[t]{7mm}{\multirow{6}{*}{\rotatebox[origin=c]{0}{OTC}}}
 & VGAE & $45.13 \pm 2.36$ & $44.97 \pm 3.10$ & $45.13 \pm 2.36$ & $44.16 \pm 2.75$ & \color{red}{$\bm{40.03 \pm 1.97}$} & \color{blue}{$45.35 \pm 2.61$}\\
 
 & SAGE & $44.48 \pm 0.67$ & $42.09 \pm 2.19$ & $44.48 \pm 0.67$ & $35.91 \pm 1.89$ & \color{red}{$\bm{0.00 \pm 0.00}$} & \color{blue}{$44.52 \pm 1.09$}\\

 & PEG & $59.56 \pm 1.05$ & $54.95 \pm 0.96$ & $59.56 \pm 1.05$ & $53.09 \pm 1.06$ & \color{red}{$\bm{52.09 \pm 1.15}$}  & \color{blue}{$59.62 \pm 1.09$}\\
 
 & GIN & $46.39 \pm 1.98$ & $45.78 \pm 1.82$ & $46.39 \pm 1.98$ & $45.18 \pm 1.59$ & \color{red}{$\bm{41.46 \pm 1.57}$} & \color{blue}{$48.67 \pm 1.47$}\\

 & GCN & $44.62 \pm 1.87$ & $42.34 \pm 1.69$ & $44.62 \pm 1.87$ & $43.92 \pm 1.69$ & \color{red}{$\bm{39.92 \pm 1.67}$} & \color{blue}{$44.98 \pm 1.98$}\\

 & SEAL & $64.32 \pm 0.57$ & $63.94 \pm 0.41$ & $64.32 \pm 0.57$ & $64.32 \pm 0.48$ & \color{red}{$\bm{62.42 \pm 0.86}$} & $64.33 \pm 0.43$\\

\midrule

 \parbox[t]{7mm}{\multirow{2}{*}{\rotatebox[origin=c]{0}{DDI}}}
 & VGAE & $37.12 \pm 0.14$ & $35.16 \pm 0.20$ & $37.12 \pm 0.14$ & $36.19 \pm 0.22$ & \color{red}{$\bm{33.26 \pm 0.29}$} & $38.97 \pm 0.37$\\
 
 & PEG & $39.07 \pm 0.20$ & $38.55 \pm 0.47$ & $39.07 \pm 0.20$ & $37.19 \pm 0.51$ & \color{red}{$\bm{34.04 \pm 0.35}$} & $39.21 \pm 0.38$\\

\midrule

 \parbox[t]{7mm}{\multirow{2}{*}{\rotatebox[origin=c]{0}{Collab}}}
 & VGAE & $31.14 \pm 0.26$ & $31.14 \pm 0.29$ & $31.14 \pm 0.26$ & $31.09 \pm 0.32$ & \color{red}{$\bm{28.15 \pm 0.34}$} & $31.14 \pm 0.25$\\
 
 & PEG & $34.22 \pm 0.30$ & $32.11 \pm 0.65$ & $34.22 \pm 0.30$ & $32.59 \pm 0.51$ & \color{red}{$\bm{30.72 \pm 0.24}$} & $34.50 \pm 0.63$\\
\bottomrule
\end{tabular}
}
\label{tab:hit20app}
\end{table*}


\end{document}